\documentclass[sigconf]{acmart}
\usepackage{booktabs} 
\usepackage{subcaption}
\usepackage{mwe}
\usepackage{graphicx}
\usepackage{tikz}
\usepackage{capt-of}
\usepackage{stackengine}

\usepackage{svg}
\usepackage{xcolor}
\usepackage{caption}
\usepackage{subcaption}
\usepackage{multirow}
\usepackage{diagbox}
\usepackage{tabularx}

\usepackage[colorinlistoftodos,prependcaption,textsize=tiny]{todonotes}

\makeatletter
\DeclareRobustCommand\onedot{\futurelet\@let@token\@onedot}
\def\@onedot{\ifx\@let@token.\else.\null\fi\xspace}

\makeatother

\newcommand{\playground}{\textsc{Play-Ground}~}
\newcommand{\fern}{\textsc{Fern}~}

\newcommand{\flower}{\textsc{Flower}~}
\newcommand{\trex}{\textsc{TRex}~}
\newcommand{\horns}{\textsc{Horns}~}

\newcommand{\stove}{\textsc{Stove}~}
\newcommand{\cuda}{\textsc{Cuda}~}

\setcopyright{rightsretained}

\acmISBN{978-1-4503-9822-0/22/12}

\acmConference[ICVGIP'22]{13th Indian Conference on Computer Vision, Graphics and Image Processing}{December 2022}{Gandhinagar, India}
\acmYear{2022}
\copyrightyear{2022}

\acmPrice{15.00}

\editor{Soma Biswas}
\editor{Shanmuganathan Raman}
\editor{Amit K Roy-Chowdhury}

\acmDOI{10.1145/3571600.3571643}

\acmArticle{43}

\copyrightyear{2022}
\acmYear{2022}
\setcopyright{acmcopyright}\acmConference[ICVGIP'22]{Proceedings of the Thirteenth Indian Conference on Computer Vision, Graphics and Image Processing}{December 8--10, 2022}{Gandhinagar, India}
\acmBooktitle{Proceedings of the Thirteenth Indian Conference on Computer Vision, Graphics and Image Processing (ICVGIP'22), December 8--10, 2022, Gandhinagar, India}
\acmPrice{15.00}
\acmDOI{10.1145/3571600.3571643}
\begin{document}
\title{StyleTRF: Stylizing Tensorial Radiance Fields}
\titlenote{\href{https://rahul-goel.github.io/StyleTRF/}{Project Page}}


\author{Rahul Goel}
\authornote{Both authors contributed equally to this research.}
\email{rahul.goel@research.iiit.ac.in}
\affiliation{%
  \institution{CVIT, KCIS, IIIT Hyderabad} \country{India}
}
\author{Sirikonda Dhawal}
\authornotemark[2]
\email{dhawal.sirikonda@research.iiit.ac.in}
\affiliation{%
  \institution{CVIT, KCIS, IIIT Hyderabad} \country{India}
}
\author{Saurabh Saini}
\email{saurabh.saini@research.iiit.ac.in}
\affiliation{%
  \institution{CVIT, KCIS, IIIT Hyderabad} \country{India}
}
\author{P. J. Narayanan}
\email{pjn@iiit.ac.in}
\affiliation{%
  \institution{CVIT, KCIS, IIIT Hyderabad} \country{India}
}

\renewcommand{\shortauthors}{Rahul G., S. Dhawal, Saurabh S. and P.J. Narayanan}
\begin{abstract}
Stylized view generation of scenes captured casually using a camera has received much attention recently. The geometry and appearance of the scene are typically captured as neural point sets or neural radiance fields in the previous work. An image stylization method is used to stylize the captured appearance by training its network jointly or iteratively with the structure capture network. The state-of-the-art SNeRF~\cite{snerf_siggraph} method trains the NeRF and stylization network in an alternating manner. These methods have high training time and require joint optimization.
In this work, we present StyleTRF, a compact, quick-to-optimize strategy for stylized view generation using TensoRF~\cite{tensorf}. The appearance part is fine-tuned using sparse stylized priors of a few views rendered using the TensoRF representation for a few iterations. Our method thus effectively decouples style-adaption from view capture and is much faster than the previous methods. We show state-of-the-art results on several scenes used for this purpose.
\end{abstract}
\begin{teaserfigure}
    \centering
    \begin{minipage}{0.3\linewidth}
        \centering
        {\includegraphics[width=\textwidth]{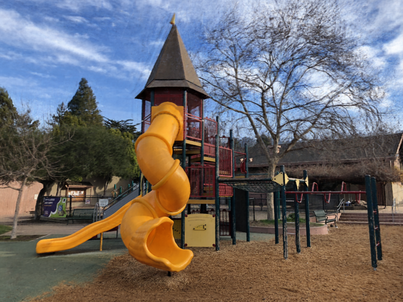}}
    \end{minipage}
    \begin{minipage}{0.3\linewidth}
        \centering
        \stackinset{r}{}{b}{}
        {\fcolorbox{black}{yellow}{\includegraphics[scale=0.11]{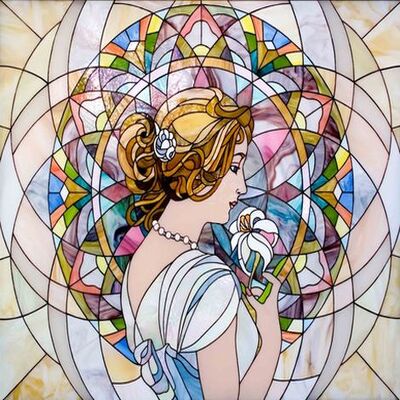}}}
        {\includegraphics[width=\textwidth]{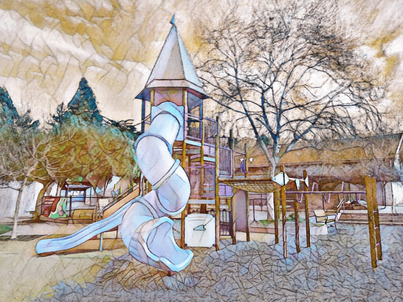}}
    \end{minipage}
    \begin{minipage}{0.3\linewidth}
        \centering
        \stackinset{r}{}{b}{}
        {\fcolorbox{black}{yellow}{\includegraphics[scale=0.08]{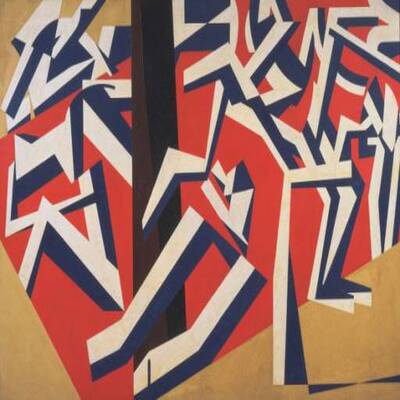}}}
        {\includegraphics[width=\textwidth]{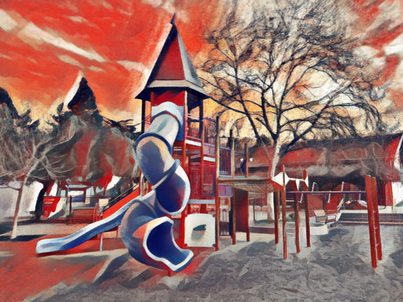}}
    \end{minipage}
     \caption{\emph{Stylization:} We show results of stylization using our technique presented in Sec. \ref{sec:method}. We stylize the \playground scene using two different styles \textit{mosaic} and \textit{mudbath}. Our Pipeline adapts style in a nominal time of 40 sec on top of  a pre-optimized TensoRF scene representation. Once the style is adapted in accordance, stylized novel views can be generated with traditional volumetric rendering techniques.}
     \label{fig:teaser}
\end{teaserfigure}


\maketitle
\begin{figure*}
    \centering
    \begin{minipage}{\linewidth}
        \centering
         \includegraphics[width=0.9\textwidth]{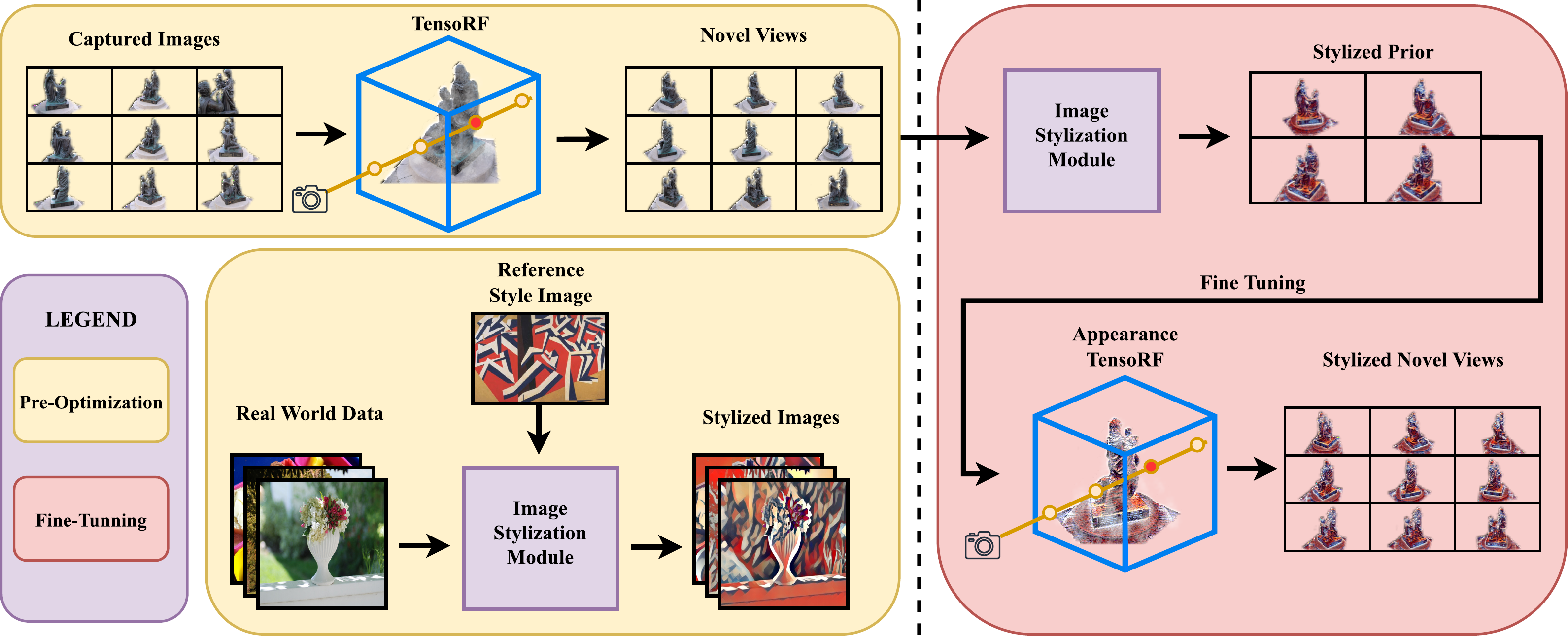}
    \end{minipage}
    \caption{\emph{System overview}: {The pipeline diagram presents an overview of the strategy employed by our work. We first optimize Tensorial Radiance Fields for representation of the scene as proposed by ~\cite{tensorf}. Concurrently we optimize stylization module utilizing ~\citet{johnson} method on COCO14-dataset\cite{coco}. The stylization module is then used to stylize a sparse set of novel views generated by the pre-optimized TensoRF. These stylized views act as sparse style-priors and are used to fine-tune the appearance of the previously optimized scene representation. It is to be noted that we freeze the density terms of the TensoRF and only alter the appearance vectors which retains geometric while adapting the novel style.}}
    \label{fig:sysDiag}
\end{figure*}

\section{Introduction}
{
    \label{sec:intro}
    Stylizing a content image based on a reference style image has been of interest to the community lately. With the development of 3D visual devices, the demand for 3D content generation has grown. Stylizing entire 3D scenes has applications in the world of augmented reality (AR) and virtual reality (VR). With style transfer on 3D scenes, one can witness an entire world through a painter's eyes.  
    
    There have been several efforts to stylize image content \cite{gatys, johnson, adain, avatar, image_style_wct, image_style_tspr}. Stylizing videos on the other hand is a harder problem as it needs temporal consistency along with stylizing across the frames of the video \cite{real_time_nst_video, ReReVST2020, ruder, video_adaattn}.
    Combining stylization with new view generation takes the game one step further. We concentrate on stylized novel view generation in this paper.

    Radiance Fields have become the dominant method of capturing a scene using a few images and then generating its views from other positions. NeRF \cite{nerf} and its successors exploit the neural representations to encode the whole scene and render novel views at test time. They use a small MLP-based architecture to represent the radiance field (typically using 8 layers of 256 neurons), which has a small memory footprint of 5MB. They, however, need very high training time, ranging between 16hrs to 1 day per scene. Plenoxels \cite{plenoxels} leverage the traditional voxel space to optimize for the feature vectors associated per voxel and rely on standard tri-linear interpolation to regress the radiance field vectors for the continuous volumetric space. The voxel-based approach benefits from the adaptive upsampling and pruning techniques, thereby reducing the optimization times to about 15-minutes. Plenoxels has a large memory footprint, often running into GBs. The recently introduced TensoRF method \cite{tensorf} uses the Tensor decomposition based on BTD and provides a way to reduce both memory and training times simultaneously. TensoRF can optimize a scene in {10-20 minutes} while maintaining a memory footprint of just {5-10MB}.
    
    Recent stylized new view generation methods use NeRF to represent the scene. They jointly or iteratively optimize NeRF with a stylization technique while rendering stylized new views \cite{style_implicit_wacv, stylizednerf_cvpr, snerf_siggraph}. The latest method SNeRF proposed by~\citet{snerf_siggraph} uses an iterative training strategy to optimize for each style using ~\citet{gatys}. Adapting to a new style is thus a heavy process, taking days of processing.
    
    In this work, we build on and modify the ideas of SNeRF \cite{snerf_siggraph} and present a simple yet powerful method for stylized new view generation. Our method nearly decouples the scene capture step from the stylization step. Each can be performed in respective pre-processing step. The joint optimization is limited to a few iterations of appearance fine-tuning that takes about 40-50 seconds to adapt to a new style. We use TensoRF \cite{tensorf} to represent the geometry and appearance of the scene, which comes with low optimization time and a small memory footprint. We use the method by \citet{johnson} for stylization. It can be trained for a particular style image (typically in under 20 minutes) ahead of time and can produce stylized images at a rate of 30 frames per sec. We generate a few novel views using the learned TensoRF representation and stylize them using Johnson's method. We then fine-tune the TensoRF's appearance branch while freezing the density components for a few iterations to obtain a stylized scene representation. Consistent, stylized novel views of the scene can be generated using this fine-tuned TensoRF representation.
    

 The main contribution of this work is using fast and light TensoRF representation for stylized view generation and devising a fast style-adaptation technique. We do this by fine-tuning the TensoRF appearance for a small number of iterations, using as prior a sparse set of images rendered from the optimized TensoRF and stylized using \citet{johnson}. This effectively decouples style-adaption from view capture and makes it much faster than methods like SNeRF.

}
\section{Related Work}
{
    \label{sec:related_work}
    \par \textbf{Stylizing Images:}
    {
        Stylizing images has been a well-studied problem in the vision community for a long time. The method proposed by \citet{gatys} optimizes white noise to match the content of one image while transferring the style from the other. \citet{johnson} proposed to use simple feed-forward architecture, which produces stylized content in real-time. While being quick at producing results, \citet{johnson} require a separate network for each style. Several works like \cite{adain, avatar, image_style_wct} have addressed this issue of style-dependent training. More recently, \citet{image_style_tspr} have been able to decompose an image into its style and content codes. While these works have provided a strong basis for stylizing 2D content, most of the aforementioned works do not faithfully stylize temporal data, let alone stylizing 3D content. 
    }
    \par \textbf{Temporal Stylization of Videos:}
    {
        The stylization of temporal-data-like videos necessitate temporal-consistent stylization. Though 2D stylization techniques provide good stylization per frame, when stitched together, they often result in flickering. This flicker is caused due to inconsistent temporal stylization. To address these issues, various methods have tried incorporating temporal-consistency losses across frames, namely \citet{ruder}. Further works like \citet{real_time_nst_video} concentrate on the real-time generation of stylization in addition to temporal-consistent stylization. Recent work in this line by \citet{ReReVST2020} has relaxed the objective function to be optimized and derived a new regularization strategy. This allowed them to stylize video content for any arbitrary given style without any re-training or fine-tuning. Though these methods provide temporally-consistent stylization, they can not produce novel stylized views. This requires scene understanding at the geometric and as well as at the appearance levels. Consequently, temporal consistency methods proposed by video stylization works cannot be extended to novel view stylization.
    }
    
    \par \textbf{Stylizing Geometric Appearance:}
    {
        As discussed earlier, obtaining stylized novel views require an understanding of the geometric content of the scene. Methods like \citet{psnet_pointcloud_style} have proposed stylizing geometry represented as 3D-point-cloud referenced on another point-cloud or image. However, stylizing point-cloud do not fully address the problem of novel view stylization as they are often accompanied by undesirable noise and missing surface information. These noise points and missing geometries can only be filled up using neural rendering-based frames works. Works like NPBG~\cite{npbg} used U-Net-based architecture to fill the missing gaps in geometry and alleviate the noisy data. Moving in this direction \citet{style_view_point_cloud} introduced consistent 3D point cloud stylization through a learned linear transformation matrix to change the appearance. They build up on NPBG~\cite{npbg} to account for the sparsity  and noisy nature of point clouds for the rendering of stylized novel views. Though the method proposed by \citet{style_view_point_cloud} produces promising stylization of novel views, the geometric representation and appearance captured by the underlying U-Net-based pipelines are not accurate.
    }
    
    \par \textbf{Novel View Synthesis(NVS) using Implicit Neural Representation:}
    {
        The Novel View Synthesis(NVS) has been a challenging field that has been tackled for decades with different approaches like Lumigraph~\cite{lumigraph}, Multi-Plane Imaging~\cite{multi_plane_images}, and Light Field Fusion~\cite{mildenhall2019llff}. Recently the field of NVS has taken a large leap forward due to the introduction of NeRF\cite{nerf}. NeRF(Neural Radiance Fields) proposed by \citet{nerf}, learns the radiance fields using simple MLPs coupled with positionally encoded input features to regress radiance at novel viewpoints, given a sparse set of posed images. Despite its great representational power, it is accompanied by limitations like slow training and rendering times, non-editable appearance, and lighting. These limitations have paved the path for various extensions ranging from speeding up its rendering process ~\cite{kilonerf,bakednerf}, edit-ability of material \cite{neuralpil, NeRD,physg2020}, relighting \cite{nerv2020,neuralpil} and improving implicit geometry \cite{neural_lumigraph}. Recently HDR based NeRFs have also garnered great interest \cite{rawnerf,hdr_gera}. Unlike these methods, we concentrate on stylizing Radiance fields with reference style images.
    }
    
    \par \textbf{Novel View Synthesis(NVS) using voxel grids:}
    {
        Apart from Implicit Neural representation of radiance fields, recently, PlenOxels~\cite{plenoxels} proposed the use of voxel grid with density and appearance associated in the form SH-vectors stored at every voxel of the grid to represent the underlying radiance field. They utilized custom \cuda implementation to achieve a training time of around 20 minutes while obtaining similar results as NeRF. However, their approach requires per-scene storage of around 1GB.
        Addressing the issue of memory footprint and requirement of custom \cuda kernels, TensoRF~\cite{tensorf} has proposed a new  VM-(Vector Matrix) Decomposition (a special case of Block Term Decomposition) \cite{btd_paper,btd_tsp}. The usage of VM-decomposition alleviated the issue of a large memory footprint while maintaining similar scene optimization times as PlenOxels.
    }
    
    \par \textbf{Stylizing Radiance Fields}
    {
        Recently, \citet{style_implicit_wacv} extended NeRF to generate stylized novel views. Similar to NeRF, they rely on simple MLPs to regress radiance fields, compositing density and radiance. To achieve stylization, they split the representation (1) volumetric density MLP and (2) appearance MLP. They use a hypernetwork~\cite{hypernetworks} trained on the style to predict the weights of the appearance MLP which introduces style information into the appearance of the scene. The losses involved in their optimization framework are memory intensive. To circumvent this issue they optimize the losses in a patch-based manner. Due to the usage of patch-based loss, their method suffers from inconsistencies across the patch borders in a single image frame, which is undesirable. Along with patch artifacts, due to different MLP splits(density and appearance), \citet{style_implicit_wacv} require large training and rendering times.
        
        Recent  work StylizedNeRF~\cite{stylizednerf_cvpr} approaches the problem of stylizing radiance fields by mutually optimizing 2D-3D consistency. For incorporating style, they replace the appearance MLP of NeRF with a style module. They also introduce consistency and mimic losses to train the style module and a 2D stylization decoder simultaneously. However, their style transfer is not accurate, and their approach involves nearly a day-long training.
        
        SNeRF~\cite{snerf_siggraph} devises a new training strategy by incrementally stylizing novel views. They first train a NeRF model on real-world data and generate novel views using it. Using the approach by \citet{gatys}, they generate a stylized version of these novel views based on a reference style. Then the NeRF model is re-trained NeRF on the so-obtained stylized views. Iteratively repeating the previous two steps stylizes a Neural radiance field representation to incorporate style information. Once the scene is stylized novel views can be generated using traditional volumetric rendering techniques. However, this learning of radiance fields involves optimization of the Neural implicit representation which comes with a heavy training cost of 4-5 days.
        
        We take inspiration from SNeRF and train a radiance field representation on stylized images themselves. However, we realize the limitations that Gatys's method introduces when generalizing the 2D images to a 3D domain and propose the usage of a different stylization module.
    }
}
\section{Method}
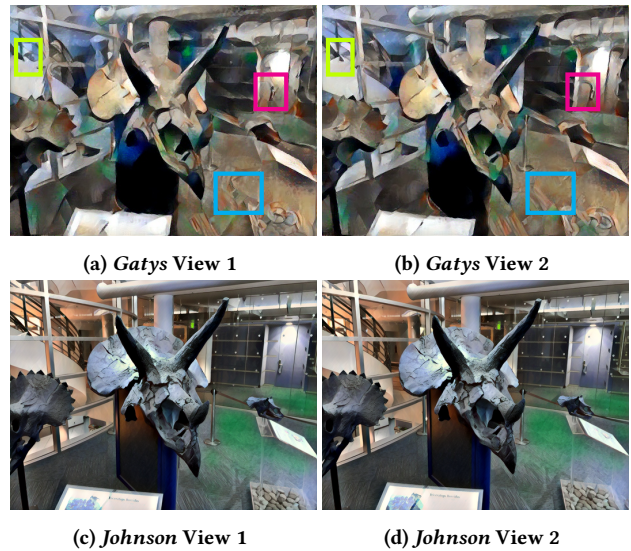
\begin{figure}
    \centering
    \begin{minipage}{0.48\linewidth}
        \centering
        \subcaptionbox{\emph{Gatys} View 1}{{
        \begin{tikzpicture}
            \node[anchor=south west,inner sep=0] (image) at (0,0) {\includegraphics[width=\textwidth]{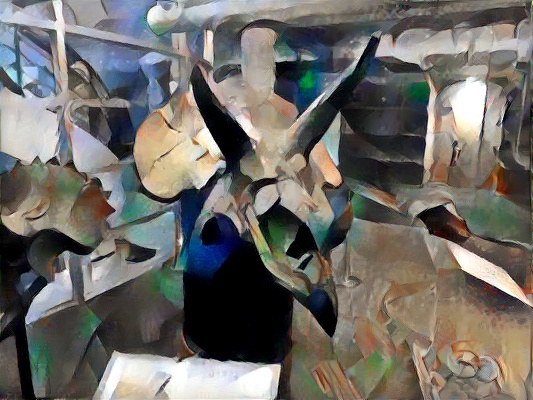}};
            \begin{scope}[x={(image.south east)},y={(image.north west)}]
                \draw[cyan,ultra thick] (0.67,0.10) rectangle (0.82,0.27);
                \draw[magenta,ultra thick] (0.8,0.55) rectangle (0.9, 0.7);
                \draw[lime,ultra thick] (0.02,0.70) rectangle (0.10, 0.85);
            \end{scope}
        \end{tikzpicture}
        }}
    \end{minipage}
    \begin{minipage}{0.48\linewidth}
        \centering
        \subcaptionbox{\emph{Gatys} View 2}{{
        \begin{tikzpicture}
            \node[anchor=south west,inner sep=0] (image) at (0,0) {\includegraphics[width=\textwidth]{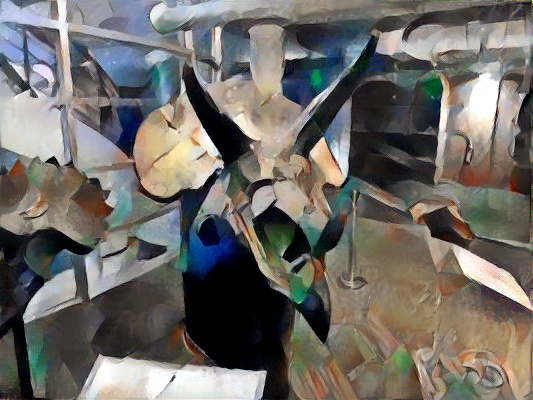}};
            \begin{scope}[x={(image.south east)},y={(image.north west)}]
                \draw[cyan,ultra thick] (0.67,0.10) rectangle (0.82,0.27);
                \draw[magenta,ultra thick] (0.8,0.55) rectangle (0.9, 0.7);
                \draw[lime,ultra thick] (0.02,0.70) rectangle (0.10, 0.85);
            \end{scope}
        \end{tikzpicture}
        }}
    \end{minipage}
    \begin{minipage}{0.48\linewidth}
        \centering
        \subcaptionbox{\emph{Johnson} View 1}{{
        \begin{tikzpicture}
            \node[anchor=south west,inner sep=0] (image) at (0,0) {\includegraphics[width=\textwidth]{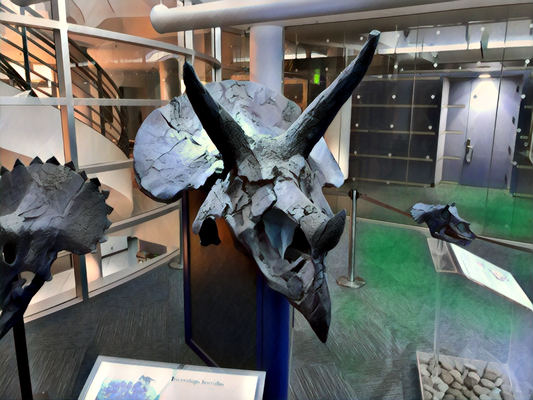}};
        \end{tikzpicture}
        }}
    \end{minipage}
    \begin{minipage}{0.48\linewidth}
        \centering
        \subcaptionbox{\emph{Johnson} View 2}{{
        \begin{tikzpicture}
            \node[anchor=south west,inner sep=0] (image) at (0,0) {\includegraphics[width=\textwidth]{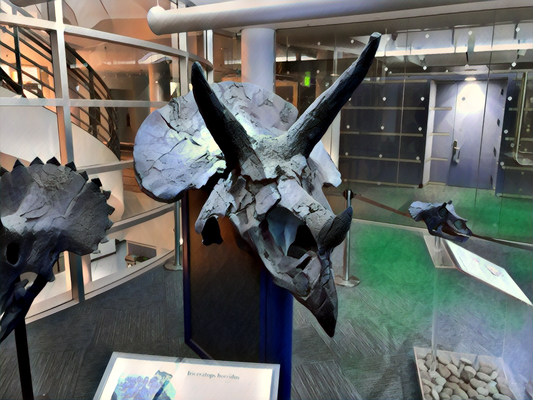}};
        \end{tikzpicture}
        }}
    \end{minipage}
    \caption{ \emph{Gatys vs Johnson Stylized Priors}: {The figures shows the output of the stylization modules proposed by \citet{gatys} and \citet{johnson} on  \horns stylized based on \textit{udnie}. \citet{gatys} produces inconsistencies in the style generated across near-by views which provide a poor prior to optimize appearance for our module. \citet{johnson} provides stable stylization across close-by views as seen in the figure.}}
    \label{fig:gatys_johnson}
\end{figure}

{
    \label{sec:method}
    Given a set of posed images of a scene and a reference style image, our aim is to render stylized novel views of the scene, which are consistent in appearance and geometry across different frames. We achieve this by fine-tuning the appearance of a TensoRF \cite{tensorf} using stylized images.
    \subsection{Preprocessing Stylization Module}
    {
        \label{sub_sec:stylization_module}
        In our method, we use the stylization method presented by \citet{johnson}, which produces stylized content in real time. The approach requires a per-style training of a CNN, for which it utilizes COCO2014 dataset~\cite{coco}. The per-style optimization takes an approximate time of 20-minutes and at the inference, it produces 30 images/sec. Since training per-style takes only 20 minutes, we can simultaneously train multiple \citet{johnson} models for each desirable style independently and infer based on the preference. 
        
        We choose \citet{johnson} as our underlying stylization module over \citet{gatys}, unlike others ~\cite{snerf_siggraph}  because it gives stable stylization for two closeby viewpoints. In the case of \citet{gatys}, stylizing two closeby viewpoints that share a large amount of image content usually results in drastically different stylization as shown in Fig.~\ref{fig:gatys_johnson}. This is because the image developed using \citet{gatys} start with white-noise and try to converge the distribution of \emph{reference-style} and \emph{content} images. This results in unstable stylization across nearby viewpoints as the optimization depends on various factors such as the learning rate, initialization, and type of optimizer. One of the reasons for the slow multi-iter stylization in SNeRF was to alleviate the inconsistencies caused by \citet{gatys} based stylization.
        
        On the other hand, once \citet{johnson} is trained, the output image is deterministic with respect to the input image. Furthermore, two close-by camera viewpoints share a large amount of spatial information. Since CNN's are spatially invariant in the local region, our image transformation network gives fairly stable stylization across closeby viewpoints Fig.~\ref{fig:gatys_johnson}. 
        
        Though~\citet{johnson} provide stable stylization across adjacent views, it is \emph{not temporally consistent}. Hence it is to be noted that, we do not rely on temporally-consistent stylization while only relying on the nominally-stable stylization of nearby viewpoints.
        The per-style training of our Stylization Module is depicted in Fig.~\ref{fig:sysDiag}.
    }
    \subsection{Scene Representation using TensoRF}
    {
        \label{sub_sec:scene_rep}
        In order to generate a novel view, it is necessary to have a geometric representation and appearance understanding. In order to address this issue, we use Radiance fields($\mathcal{L}$). The radiance fields~\cite{source_radiance_fields} ($\mathcal{L}$) is given by:
        \begin{equation}
            \label{eq:radiance_fields}
            \mathcal{L}:R^3 \times S^2 \rightarrow R^3
        \end{equation}
        where $R^3$ on the left is the scene’s world space; $S^2$ is the sphere of directions about each point, and the $R^3$ on the right is radiance at the point. Radiance fields in a way encode geometry and radiance in their mapped representation.
        Though there exist various recent adaptions of the mapping function presented in Eq.(\ref{eq:radiance_fields}) like NeRF\cite{nerf}, PlenOxel\cite{plenoxels}, we chose TensoRF~\cite{tensorf} which is a compact, accurate, and fast-to-optimize representation of Radiance Fields. We use a specifically \emph{VM-48} variant of TensoRF which optimizes a scene within a timeframe of 15-20 minutes while maintaining a memory footprint of 10-15MB.

        The TensoRF representation utilizes a Tensor decomposition known as VM-decomposition which in itself is a special case of BT-decomposition. This reduces the voxel grid memory by order of $\mathcal{O}(n)$. This scene optimization can be done independently for every scene irrespective of style. In a later phase, we alter the appearance to adapt to the reference style in a short period of 40-50 seconds.

        Similar to TensoRF~\cite{tensorf}, we utilize $L1$ norm loss and total variation (TV) loss (Eq. \ref{eq:tv_loss}) for regularization. This helps our process to avoid getting stuck in local minima and prevents overfitting. For scenes with less number of captured images, TV loss is a better choice to obtain good results. The equation for TV loss is given by:

        \begin{equation}
            \label{eq:tv_loss}
            \mathcal{L}_{TV} = \dfrac{1}{N} \sum \left( \sqrt{\Delta ^2 T_{\sigma}} + 0.1  \sqrt{\Delta ^2 T_{C}} \right)
        \end{equation}

        Here, $\Delta ^2$ is the squared difference between the neighboring values in our tensors, $N$ is the total number of parameters across our TensoRF representation $T$. $T_{\sigma}$ represents the density value and $T_{C}$ represent the appearance value in the TensoRF representation. They are weighted in the ratio of $10:1$ respectively. More details about TV Loss can be found in the work by \citet{tensorf}.
    }
    \subsection{Stylizing TensoRF representation}
    {
        \label{sub_sec:stylizing_tensorf}
        \subsubsection{Novel View stylization}
        {
            \label{subsub_sec:novel_view_style}
            Upon optimizing the radiance fields which encode the geometry and radiance of the scene as discussed in the Sec.\ref{sub_sec:scene_rep}, we render a sparse set of $20-30$ novel views in a simple trajectory (spiral). We stylize these novel renders using the pre-trained Stylization Module discussed in \ref{sub_sec:stylization_module}. Fig.~\ref{fig:sysDiag} (top-right) shows the generation stylization of these renders utilizing the Stylization Module.
        }
        \subsubsection{Stylizing Appearance of TensoRF}
        {
            \label{subsub_sec:appearance_style}
            We utilize the sparse set of stylized novel views generated using the per-style optimized \citet{johnson} module and optimize the appearance vectors of the TensoRF. During the process of optimization, we ensure that the density terms are frozen, and only the appearance is altered. We explicitly chose to freeze density as we have observed that stylizing looks pleasing and free from artifacts when density is kept frozen.  This fine-tuning only takes a nominal time of downwards of 40 secs. Once the fine-tuning is done, we obtain a geometric scene represented as Tensorial Radiance fields, which can be used to render \emph{stylized novel views} with consistent appearance across the viewpoints. The rendering of each image having a resolution of $800 \times 800$ takes an approximate time of 4-5 seconds. Fig.~\ref{fig:sysDiag} (bottom-right) shows the appearance modification from a sparse set of inputs and novel view generation.
        }
    }

}

\section{Implementation Details}
{
    \label{sec:implementation}
    \subsection{Optimizing TensoRF}
    {
        \label{sub_sec:implementation_tensorf}
        The training/optimization of radiance fields requires information of the camera poses from which an image is captured. In the case of real scenes, we rely on COLMAP~\cite{colmap}to obtain this information and in the case of synthetic scenes, we use the data obtained from Blender. We optimize TensoRF (VM-Split-48) on the input images for $15k$ iterations. In each iteration, we shoot $4096$ rays into the voxel grid. We obtain the radiance using volumetric rendering (Eq. \ref{eq:volumetric}) and optimize the grid iteratively.
        \begin{equation}
            \label{eq:volumetric}
            \mathcal{C} = \sum_{q=1}^{Q} \tau_q \left(1 - exp\left(-\sigma_q  \Delta_q \right) \right) c_q,\ \tau_q = exp\left(-\sum_{p=1}^{q-1}\sigma_p \Delta_p \right)
        \end{equation}
        We use Adam optimizer \cite{adam} which is initialized to a learning rate of $0.02$ and is re-initialized to $0.02$ after upsampling. The voxel grid is initialized with an effective resolution of $128^3$ and iteratively upsampled every $1000$ iterations,  first upsampling starting $2000$ until $5000$ iterations are reached. We finally reach an effective voxel-grid resolution of $640^3$ for real-world scenes and $300^3$ for synthetic scenes. It is to be noted that the resolution mentioned here is \emph{effective} but not exact, as the VM-decomposition provided by TensoRF presents a compact representation of voxel grid. Similar to TensoRF, we bi-linearly interpolate the matrix and linearly interpolate the vector in the VM decomposed representation during upsampling.
        This (bi-linear + linear) interpolation is similar to the tri-linear interpolation of the \emph{full-voxel} grid. We perform such interpolations with neighboring voxels during the evaluation of a ray query. This enables us to obtain continuity in our rendered images. This pre-optimization phase requires 10-15 minutes. Once optimized it can be used to generate new views with various camera poses which post-stylization act as priors to our fine-tuning phase.

    }
    \subsection{Stylization}
    {
        \label{sub_sec:implementation_stylization}
        Independently, we train the stylization module \citet{johnson} with the various reference style image on the COCO-14 Dataset \cite{coco} which takes 20 minutes. We train multiple such \citet{johnson} models for each desired style as the time required to train is quite less. We use this to create stylized priors from the views generated in the previous phase.

        While optimizing for the style we freeze the density parameters in the TensoRF representation and optimize only the appearance parameters for a small number of iterations($1k$ iterations) with the stylized prior. This style adaption only takes a nominal time of 40-50 seconds. The style-adapted TensoRF representation obtained in the previous step can be thus used to generate novel stylized views using traditional voxel rendering techniques. The generation of each view takes around 4-5 seconds.
    }
    
}
\vspace{-3mm}
\begin{figure}
\includegraphics[width=\linewidth]{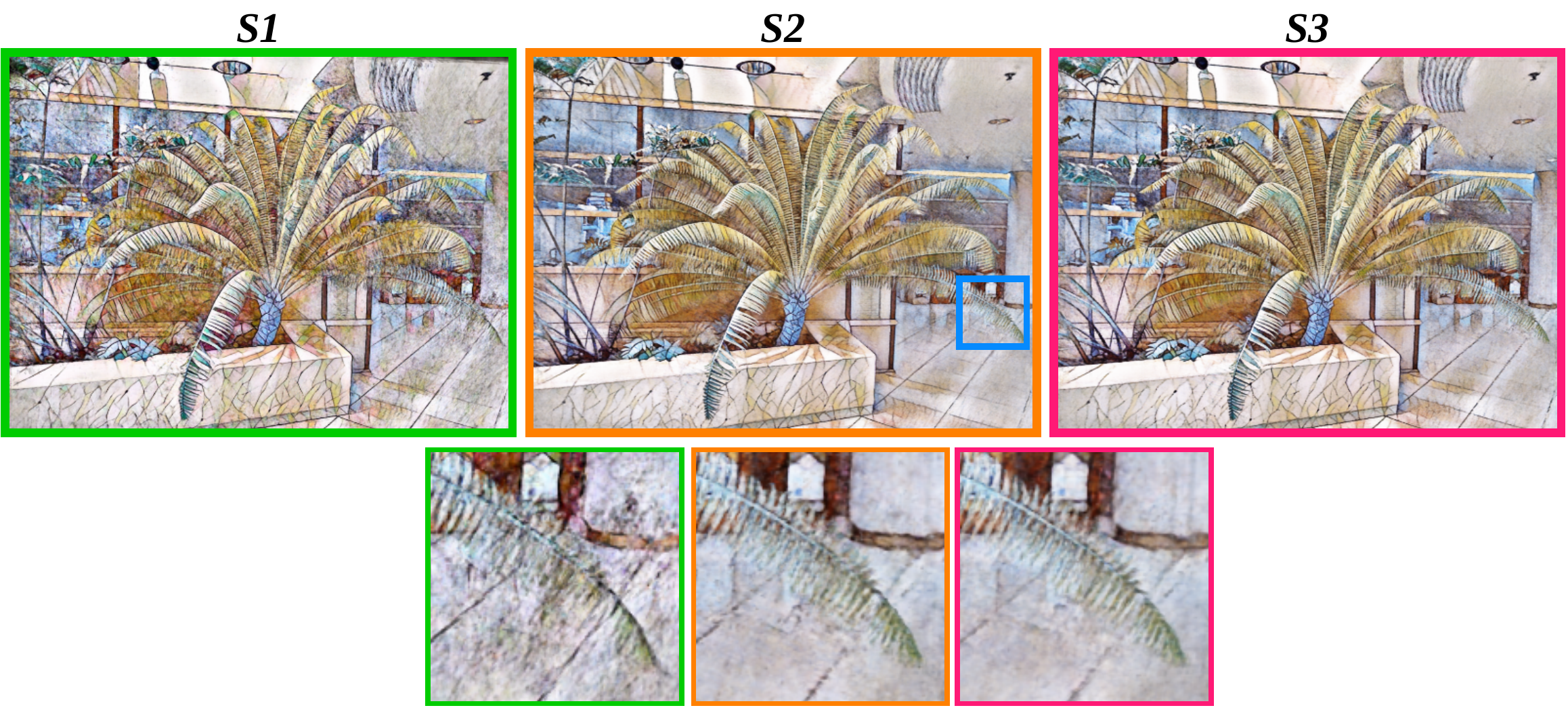}
\caption{ \emph{Comparison of the different optimization strategies:} {
Strategy \emph{S1} optimized directly on stylized images produces massive artifacts due to the loss in geometry. Unconstrained optimization \emph{S2} on the other hand achieves results compared to \emph{S1} while struggling to capture micro-details as seen in the insets. Our StyleTRF on the other hand freezes density while only optimizing for appearance, capturing the style while maintaining the geometric detail of the original scene.
}}
\label{fig:fulltrained_pretrained}
\end{figure}

\section{Experiments}
\begin{figure*}
    \centering
    \begin{minipage}{0.24\linewidth}
    \centering
    \stackinset{l}{}{b}{}
    {\fcolorbox{black}{green}{\includegraphics[scale=0.15]{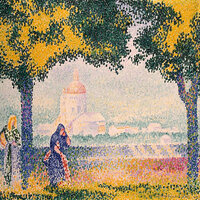}}}
    {\includegraphics[width=\textwidth]{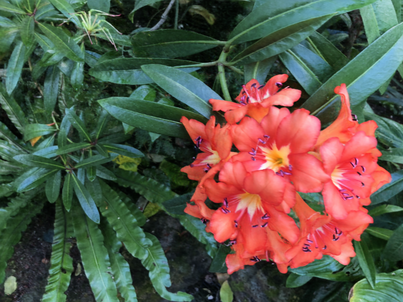}}
    \end{minipage}
    \begin{minipage}{0.24\linewidth}
        \centering
         \includegraphics[width=\textwidth]{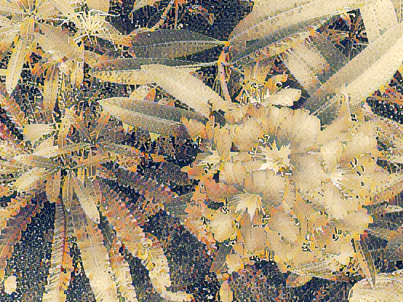}
    \end{minipage}
    \begin{minipage}{0.24\linewidth}
        \centering
         \includegraphics[width=\textwidth]{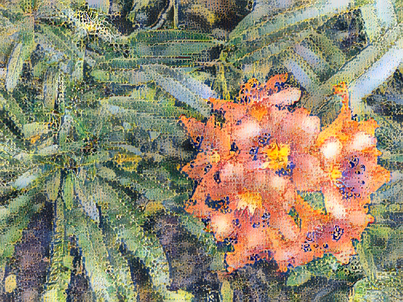}
    \end{minipage}
    \begin{minipage}{0.24\linewidth}
        \centering
         \includegraphics[width=\textwidth]{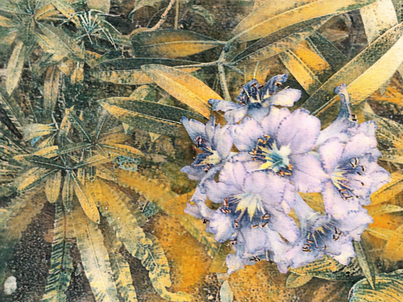}
    \end{minipage}
    \begin{minipage}{0.24\linewidth}
        \centering
        \stackinset{l}{}{b}{}
        {\fcolorbox{black}{cyan}{\includegraphics[scale=0.15]{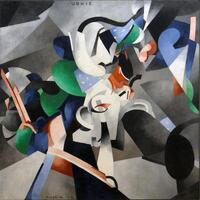}}}
        {\includegraphics[width=\textwidth]{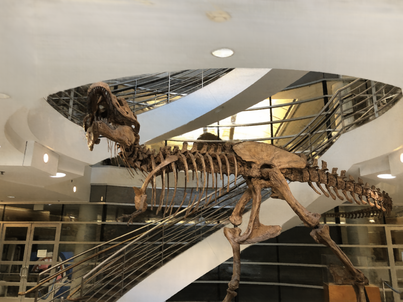}}
    \end{minipage}
    \begin{minipage}{0.24\linewidth}
        \centering
         \includegraphics[width=\textwidth]{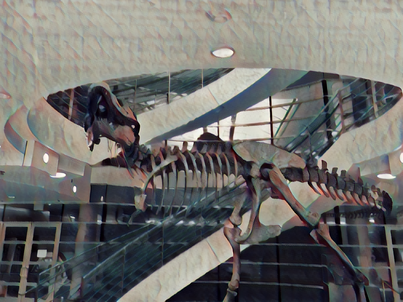}
    \end{minipage}
    \begin{minipage}{0.24\linewidth}
        \centering
         \includegraphics[width=\textwidth]{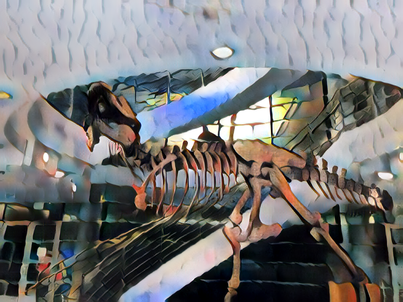}
    \end{minipage}
    \begin{minipage}{0.24\linewidth}
        \centering
         \includegraphics[width=\textwidth]{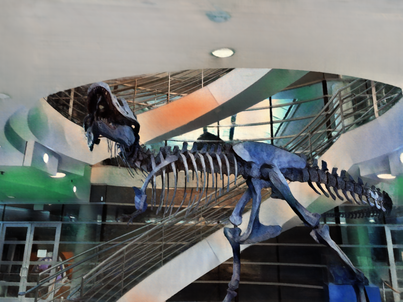}
    \end{minipage}
    \begin{minipage}{0.24\linewidth}
        \centering
        \subcaptionbox{GT Unstylized}{{\stackinset{l}{}{b}{}
        {\fcolorbox{black}{orange}{\includegraphics[scale=0.65]{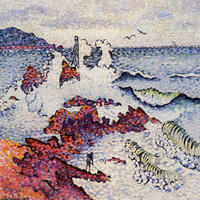}}}
        {\includegraphics[width=\textwidth]{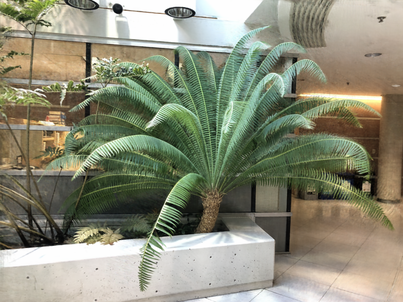}}}}
    \end{minipage}
    \begin{minipage}{0.24\linewidth}
        \centering
         \subcaptionbox{ReReVST}{{\includegraphics[width=\textwidth]{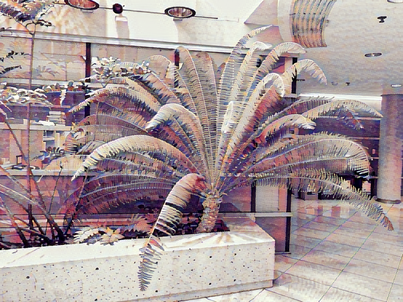}}}
    \end{minipage}
    \begin{minipage}{0.24\linewidth}
        \centering
         \subcaptionbox{StylizedNeRF}{{\includegraphics[width=\textwidth]{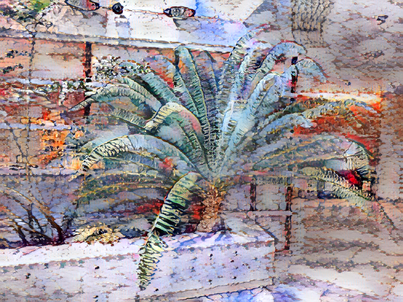}}}
    \end{minipage}
    \begin{minipage}{0.24\linewidth}
        \centering
         \subcaptionbox{StyleTRF (Ours)}{{\includegraphics[width=\textwidth]{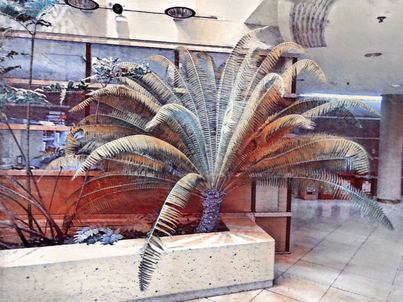}}}
    \end{minipage}
    \caption{ \emph{Qualitative Comparison}:
    {[Row 1: \textit{santamaria}, Row 2: \textit{udnie}, Row 3: \textit{mediterranean}]}
    {Here we show the comparisons Un-stylized frame in column-(a), temporally consistent stylization of ReReVST~\cite{ReReVST2020}in column-(b),  StylizedNeRF~\cite{stylizednerf_cvpr} in column-(c) and stylization using our pipeline in column-(d). It can be observed that due to the usage of combined neural representation density and radiance, the style adaptation is affecting in the case of StylizedNeRF. Observe noisy geometric structures in \textit{mediter} applied onto \fern}}
    \label{fig:results}
\end{figure*}

\begin{figure}[b]
    \centering
    \begin{minipage}{0.49\linewidth}
        \centering
         \subcaptionbox{5 Priors}{\includegraphics[width=\textwidth]{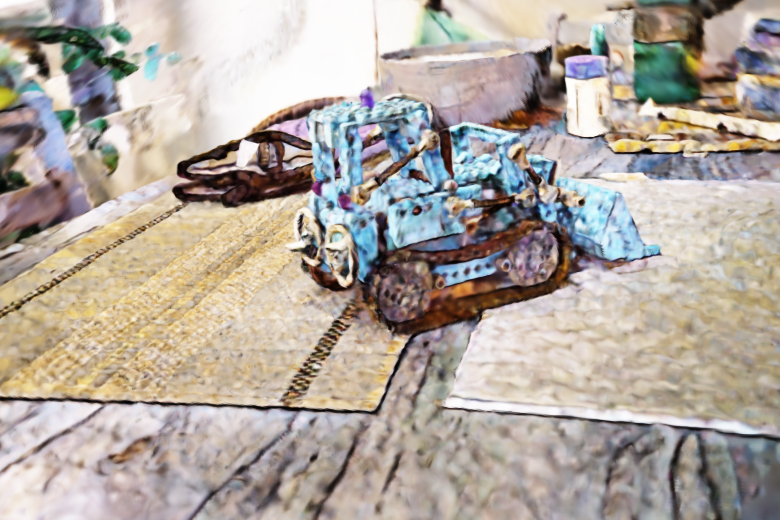}}
    \end{minipage}
    \begin{minipage}{0.49\linewidth}
        \centering
         \subcaptionbox{10 Priors}{\includegraphics[width=\textwidth]{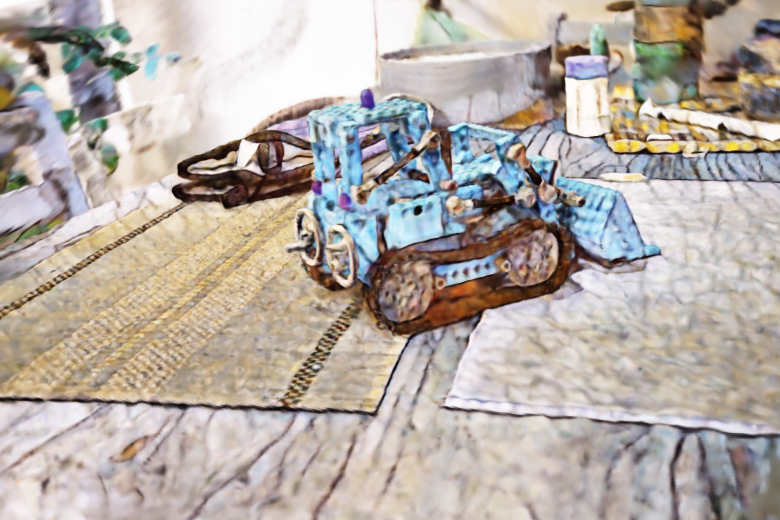}}
    \end{minipage}
    \begin{minipage}{0.49\linewidth}
        \centering
         \subcaptionbox{20 Priors}{\includegraphics[width=\textwidth]{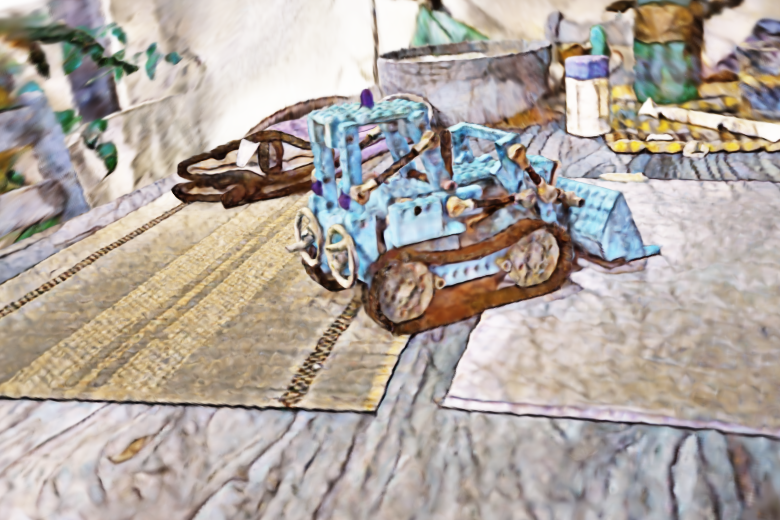}}
    \end{minipage}
    \begin{minipage}{0.49\linewidth}
        \centering
         \subcaptionbox{40 Priors}{\includegraphics[width=\textwidth]{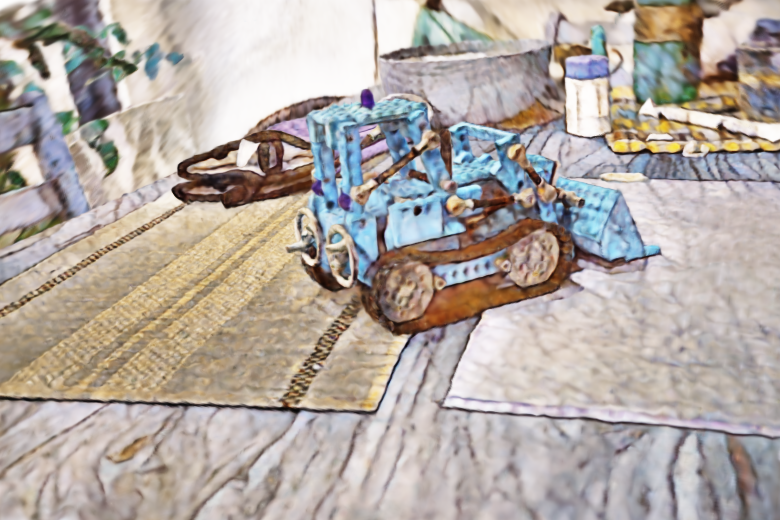}}
    \end{minipage}
    \caption{ \textcolor{black}{\emph{Number of Stylized Priors}:
    {The images in this figure are obtained by fine-tuning StyleTRF using a different number of priors on the kitchen scene from the MipNeRF360 \cite{mipnerf360} dataset which is an \textbf{unbounded 360 degree} scene. Depending on priors quality improves and saturates at 35-40 priors.}}}
    \label{fig:unboun_360}
\end{figure}

{
 \label{sec:experiments}
  In this section, we present the various experiments with the proposed StyleTRF.
  We conduct all the experiments mentioned throughout the paper including the comparisons, on a workstation PC equipped with an AMD Ryzen-5800x and an NVidia RTX-3090 GPU. In Sec.~\ref{sub_sec:exp_stylization_module} we discuss the choice of stylization module used to stylize the sparse prior used in our approach. We also experimentally compare and contrast between temporal-stylization of smooth trajectories of ground truth 3D content vs Actual 3D-Stylization in Sec.~\ref{sub_sec:video_vs_3d}. Finally, in Sec.~\ref{sub_sec:optimization_ablation} we show the effect of different optimization strategies used to stylize the 3D content.
  
 \subsection{Stylization Module}
 {
  \label{sub_sec:exp_stylization_module}
  For stylizing the sparse prior required by our method we use \citet{johnson}. The concurrent work SNeRF\cite{snerf_siggraph} uses \citet{gatys} method to iteratively stylize the radiance fields. We choose \citet{johnson} over \citet{gatys} because the latter depends on several factors such as the initialization, learning rate, and the optimization method which make it unreliable to obtain stable stylization across close-by views let alone temporal consistency. This can be observed in Fig.~\ref{fig:gatys_johnson}. This is one of the reasons, \citet{snerf_siggraph} requires multiple epochs to reflect the style in appearance.
  
  On the other hand, \citet{johnson} use a fixed CNN-based architecture to infer the stylized image. Due to the spatial consistency of CNNs, two close-by views sharing significant spatial content lead to stable stylized close-by views.
 }
 
 \subsection{Video Stylization v/s 3D Stylization}
 {
  \label{sub_sec:video_vs_3d}
  It can be argued that instead of stylizing 3D content, one can generate novel views on a camera trajectory and use temporally consistent stylization frameworks like ReReVST \cite{ReReVST2020} to obtain stylized novel views.
  
  However, we have observed that though temporal stylization is maintained, the work of ReReVST fails to fully capture the style information. The same can not be said for the stylization of 3D content. In Fig.~\ref{fig:results} it can be seen that both StylizedNeRF and our method capture better style compared to ReReVST.
 }
 \subsection{Optimization Strategies for Style Adaptation}
 {
    \label{sub_sec:optimization_ablation}
    For the stylization of Radiance fields we present three strategies:
    \begin{enumerate}
        \item \emph{S1: } Optimizing a TensoRF from scratch directly using the stylized priors.
        \item \emph{S2: } Pre-optimizing a TensoRF on the original ground truth and adapting for style using sparse stylized priors \emph{without freezing any parameters (both density and appearance)}.
        \item \emph{S3: } Pre-optimizing a TensoRF on the original ground truth data and adapting for style using sparse stylized priors while freezing the density.
    \end{enumerate}
    When following the \emph{S1} we observed geometric artifacts. This is because the stylized priors generated may not share the exact geometry with the ground truth. As stated above in Sec.~\ref{sub_sec:stylization_module}, \citet{johnson} does not provide temporally consistent stylization which might also affect the geometry so-optimized in the stylized views in the case of \emph{S1}.
    
    Another strategy \emph{S2} produces considerably better results compared to \emph{S1} as seen in Fig.~\ref{fig:fulltrained_pretrained}. This is because most of the geometric prior is learned from the pre-optimization phase which uses ground-truth images to obtain scene properties. 

    Though \emph{S2} has produced good stylization at a micro-level, fuzzy geometry can be observed which reduces the appeal of the stylization. Specifically, thin geometric structures suffer from these undesirable fuzzy geometric changes. The limit-free optimization strategy of \emph{S2} fiddles with the geometry components and leads to these artifacts. This can be observed in the insets provided in Fig.~\ref{fig:fulltrained_pretrained}.

    Our StyleTRF (\emph{S3}) approach on the other hand alleviates these issues by freezing the geometric components of scene representation. Our approach generates crispier results compared to the rest of the aforementioned strategies \emph{S1, S2}. This behavior is consistent across all the scenes.
    
    \textcolor{black}{We experiment with a different number of stylized priors to stylize the underlying scene. As shown in Fig. \ref{fig:unboun_360}, we find that good stylization is obtained when all the priors cover the entirety of the scene. In most cases, 30-40 randomly sampled camera positions around the object suffice.}
 }
 
}

\begin{table*}[t]
        \centering
        \begin{tabular}{|c|c|c|c|c|c|c|}\hline
            \multirow{2}{*}{\backslashbox{Scene}{Method}} & \multicolumn{2}{c|}{\textit{ReReVST\cite{ReReVST2020}}} & \multicolumn{2}{c|}{\textit{StylizedNeRF\cite{stylizednerf_cvpr}}} & \multicolumn{2}{c|}{\textit{Ours}} \\
            \cline{2-7}
                                                          &  \textit{short-term}   & \textit{long-term}   & \textit{short-term}  & \textit{long-term} & \textit{short-term}     &    \textit{long-term} \\
            \hline
            \textit{\horns}                               &  0.0046   & 0.0137   & 0.0229 & 0.0239 & \textbf{0.0040} & \textbf{0.0120} \\
            \textit{\fern}                                &  0.0028   & 0.0080   & 0.0100 & 0.0168 & \textbf{0.0020} & \textbf{0.0069} \\
            \textit{\flower}                              &  0.0039   & 0.0106   & \textbf{0.0020} & 0.0277 & 0.0030 & \textbf{0.0089} \\
            \hline
            
        \end{tabular}
        \caption{\emph{Consistency Metrics:} We show \textit{short-term} and \textit{long-term} consistency metrics across a smooth trajectory generated using our appearance stylized scene representation. We have found that we obtain better \textit{short \& long-term} consistency compared to the SOTA 3D-Stylization technique StylizedNeRF\cite{stylizednerf_cvpr}, while maintaining better style transfer compared to temporal stylizing methods like ReReVST\cite{ReReVST2020}.}
        \label{tab:consistency_metrics}
\end{table*}
\begin{figure*}
    \centering
    \begin{minipage}{0.24\linewidth}
        \centering
        \begin{tikzpicture}
            \node[anchor=south west,inner sep=0] (image) at (0,0)        {\includegraphics[width=\textwidth]{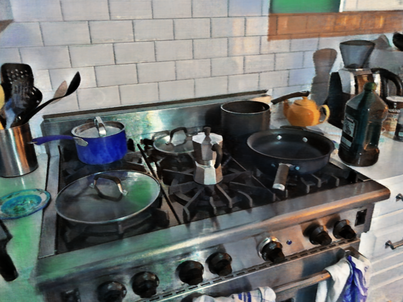}};
            \begin{scope}[x={(image.south east)},y={(image.north west)}]
                \draw[cyan,ultra thick] (0.25,0.25) rectangle (0.45,0.45);
                \draw[magenta,ultra thick] (0.8,0.0) rectangle (1.0,0.25);
            \end{scope}
        \end{tikzpicture}
    \end{minipage}
    \begin{minipage}{0.24\linewidth}
        \centering
        \begin{tikzpicture}
            \node[anchor=south west,inner sep=0] (image) at (0,0)        {\includegraphics[width=\textwidth]{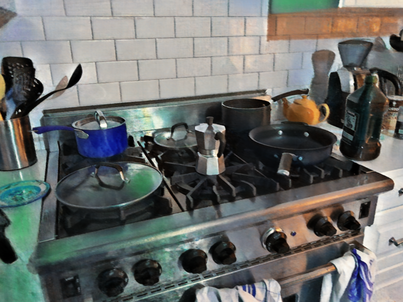}};
            \begin{scope}[x={(image.south east)},y={(image.north west)}]
                \draw[cyan,ultra thick] (0.25,0.25) rectangle (0.45,0.45);
                \draw[magenta,ultra thick] (0.8,0.0) rectangle (1.0,0.25);
            \end{scope}
        \end{tikzpicture}
    \end{minipage}
    \begin{minipage}{0.24\linewidth}
        \centering
        \begin{tikzpicture}
            \node[anchor=south west,inner sep=0] (image) at (0,0)        {\includegraphics[width=\textwidth]{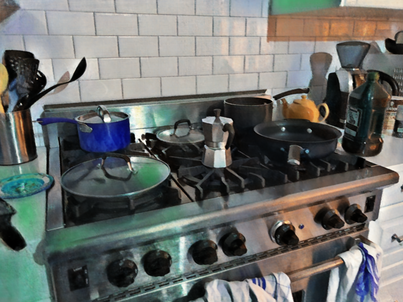}};
            \begin{scope}[x={(image.south east)},y={(image.north west)}]
                \draw[cyan,ultra thick] (0.3,0.3) rectangle (0.5,0.5);
                \draw[magenta,ultra thick] (0.8,0.0) rectangle (1.0,0.25);
            \end{scope}
        \end{tikzpicture}
    \end{minipage}
    \begin{minipage}{0.24\linewidth}
        \centering
        \begin{tikzpicture}
            \node[anchor=south west,inner sep=0] (image) at (0,0)        {\includegraphics[width=\textwidth]{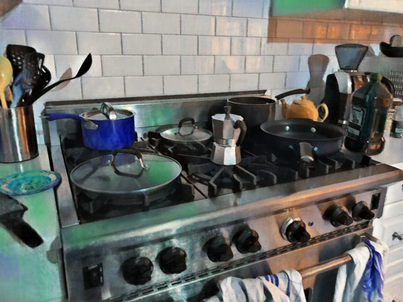}};
            \begin{scope}[x={(image.south east)},y={(image.north west)}]
                \draw[cyan,ultra thick] (0.3,0.3) rectangle (0.5,0.5);
                \draw[magenta,ultra thick] (0.8,0.0) rectangle (1.0,0.25);
            \end{scope}
        \end{tikzpicture}
    \end{minipage}

    \begin{minipage}{0.24\linewidth}
        \centering
        \begin{tikzpicture}
            \node[anchor=south west,inner sep=0] (image) at (0,0)        {\includegraphics[width=\textwidth]{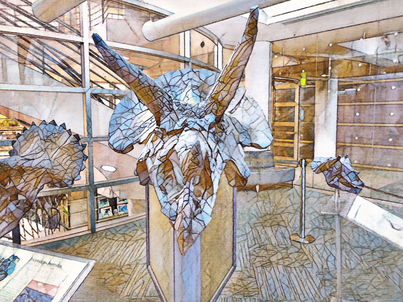}};
            \begin{scope}[x={(image.south east)},y={(image.north west)}]
                \draw[blue,ultra thick] (0.05, 0.4) rectangle (0.25,0.6);
                \draw[magenta,ultra thick] (0.35,0.25) rectangle (0.6,0.5);
            \end{scope}
        \end{tikzpicture}
    \end{minipage}
    \begin{minipage}{0.24\linewidth}
        \centering
        \begin{tikzpicture}
            \node[anchor=south west,inner sep=0] (image) at (0,0)        {\includegraphics[width=\textwidth]{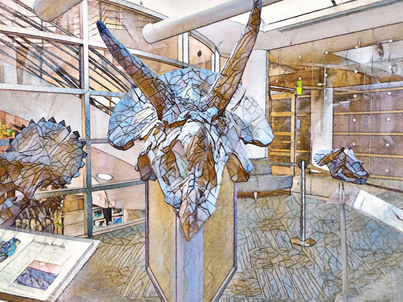}};
            \begin{scope}[x={(image.south east)},y={(image.north west)}]
                \draw[blue,ultra thick] (0.05, 0.4) rectangle (0.25,0.6);
                \draw[magenta,ultra thick] (0.35,0.25) rectangle (0.6,0.5);
            \end{scope}
        \end{tikzpicture}
    \end{minipage}
    \begin{minipage}{0.24\linewidth}
        \centering
        \begin{tikzpicture}
            \node[anchor=south west,inner sep=0] (image) at (0,0)        {\includegraphics[width=\textwidth]{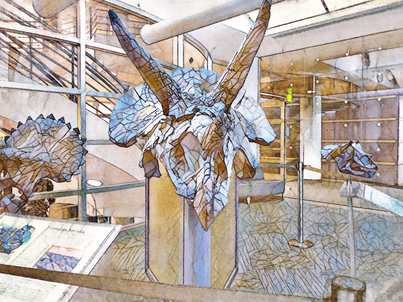}};
            \begin{scope}[x={(image.south east)},y={(image.north west)}]
                \draw[blue,ultra thick] (0.05, 0.4) rectangle (0.25,0.6);
                \draw[magenta,ultra thick] (0.40,0.30) rectangle (0.65,0.55);
            \end{scope}
        \end{tikzpicture}
    \end{minipage}
    \begin{minipage}{0.24\linewidth}
        \centering
        \begin{tikzpicture}
            \node[anchor=south west,inner sep=0] (image) at (0,0)        {\includegraphics[width=\textwidth]{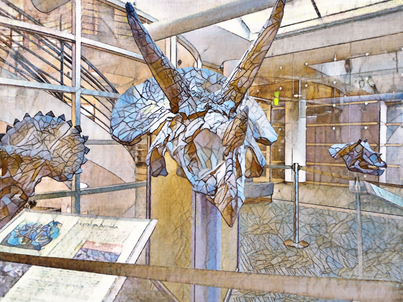}};
            \begin{scope}[x={(image.south east)},y={(image.north west)}]
                \draw[blue,ultra thick] (0.05, 0.4) rectangle (0.25,0.6); 
                \draw[magenta,ultra thick] (0.40,0.30) rectangle (0.65,0.55);
            \end{scope}
        \end{tikzpicture}
    \end{minipage}
    \caption{ \emph{View Consistency Across Frames}: {The figure shows stylized novel views of a simple trajectory. To keep it simple we named the frames in $t_i$ with increasing order of $i$ from left to right. It can be observed clearly that our stylization is multi-view consistent both in the case of \stove and \horns.}}
        \label{fig:view_consistency}
\end{figure*}
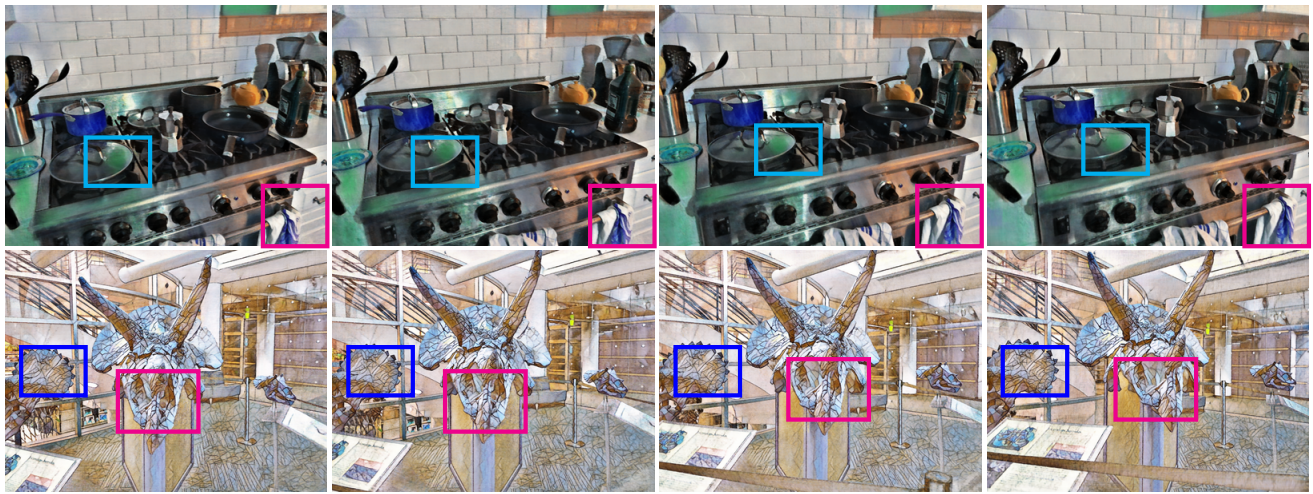

\section{Results}
{
    \label{sec:results}
    In this section, we compare our results against both 3D-stylization techniques and temporally-consistent stylization techniques both quantitatively and qualitatively. 
    \subsection{Qualitative Results}
    {
        \label{sub_sec:qualitative_results}
        For comparing qualitative results, we can use smooth trajectories which are similar to a video. This similarity enables us to compare with temporally-consistent stylization techniques alongside 3D-Stylization. For 3D stylized novel view synthesis, though there exists \citet{style_view_point_cloud}, they do not hold an accurate representation of geometry and rely on explicit geometry as input. Although other methods like \citet{style_implicit_wacv} do not rely on explicit geometry input, they suffer from patch border artifacts as discussed in ~\cite{stylizednerf_cvpr} and Sec. \ref{sec:related_work}. Hence, we compare our results with the latest 3D radiance field-based stylization method StylizedNeRF~\cite{stylizednerf_cvpr} and temporally-consistent stylization technique of ReReVST~\citet{ReReVST2020}, on similar lines as \cite{stylizednerf_cvpr}. 
        \par{\textbf{Comparison amongst different stylization Techniques:}}
        {
            We have observed that though the novel views generated using StylizedNeRF are reasonably consistent,  the geometry after stylization in the case of StylizedNeRF has been greatly affected as seen in Fig.~\ref{fig:results}(c). It produces blurry geometry in the case of \trex stylized using \textit{udnie} along with the missing greenish tints present in the \textit{udnie}, and produces extremely noisy results for \textit{mediterranean} applied on \fern. Contrary to this, our method distinctively transfers different colors to different parts of the scene as seen in \trex stylized using \textit{udnie} and captures the complete color palette in \textit{mediterranean} adapted onto \fern. Our method also consistently preserves geometry across all scenes and styles. The geometric noise in the StylizedNeRF can partly be attributed to the combined density and appearance encoded of a single MLP, which hinders the disentanglement of geometry and radiance as observed by \cite{snerf_siggraph}. 
            
            In the case of the temporal-consistent style transfer technique ReReVST, we have observed that although the method is robust to adapt to new styles on the fly and produces results in real-time, it lacks proper capture of style information. This can be observed Fig.~\ref{fig:results}(b), ReReVST fails to capture the prevalent blue tint in \textit{santamaria} in the case of \flower, this could also be observed in the case of StylizedNeRF and vaguely captures the color palette of \textit{mediterranean} in \fern. For \textit{udnie} applied on \trex, it does not account for distinct colors present in the style and the resultant image contains an unappealing blend of colors throughout the image.
        }
        \par{\textbf{View Consistency:}}
        {
            To qualitatively show view consistency, we render views across a smooth trajectory and show different views across it. In Fig.~\ref{fig:view_consistency} we show our renders for \textit{udnie} and \textit{mosiac} styles adapted onto \stove and \horns respectively. It can be observed from the insets of Fig.~\ref{fig:view_consistency} that stylized radiance across the frame is consistent, validating our claim that 3D-stylization-based novel view generation can produce multi-view consistent stylized content.
        }
    }
    \subsection{Quantitative Results}
    {
        \label{sub_sec:quantitative_results}
        In order to check the consistency of our stylized content, we generate a smooth trajectory similar  to SNeRF~\cite{snerf_siggraph}. The rendered views along the trajectory are used to evaluate the consistency. Since the smooth trajectory replicates the behavior of a video we can comfortably compare our method with the temporal-consistent stylization techniques like ReReVST\cite{ReReVST2020} alongside 3D-stylization methods like StylizedNeRF\cite{stylizednerf_cvpr}. For this comparison, we chose \fern, \flower, and \horns scenes and report the respective metrics in Tbl.~\ref{tab:consistency_metrics}. We obtained better metrics compared to both temporal stylization~\cite{ReReVST2020} and implicit geometric stylization~\cite{stylizednerf_cvpr}. We estimated the consistency by calculating the optical flow $\smash{\mathcal{O}}$ between the non-stylized frames $\smash{\mathcal{F}^{real}_i}$ and $\smash{\mathcal{F}^{real}_{i + \delta}}$ rendered using TensoRF. 
        \begin{equation}
            \label{eq:optical_flow}
            \mathcal{O}:opticalflow\left(\mathcal{F}^{real}_{i}, \mathcal{F}^{real}_{i + \delta}\right)
        \end{equation} 
        Using the so-obtained optical-flow $\smash{O}$, we warp the stylized frame $\smash{\mathcal{F}^{style}_i}$ to $\smash{\hat{ \mathcal{F}}^{style}_{i + \delta}}$. 
        
        \begin{equation}
            \label{eq:warp}
            \mathcal{W}:\hat{\mathcal{F}}^{style}_{i + \delta} \leftarrow Warp(\mathcal{F}^{style}_{i}, \mathcal{O})
        \end{equation}
        
        For the calculation of optical flow, we use RAFT~\cite{raft} similar to SNeRF. 
        
        We then calculate the pixel-averaged $L_2$ loss between the warped frame ${\hat{ \mathcal{F}}^{style}_{i + \delta}}$ and the actual stylized frame ${\mathcal{F}^{style}_{i + \delta}}$. We aggregate this loss across the frame-combinations and report the metrics in Tbl.~\ref{tab:consistency_metrics}. For calculation of \textit{short-term} consistency we chose $\delta = 1$ and for \textit{long-term} $\delta = 5$. It is to be noted that $\delta$ here represents the change in the camera position along the trajectory. We have observed that though the \textit{temporal-cosistent} stylization techniques like \citet{ReReVST2020} come close to our results, they struggle to capture the reference-style (as seen in Fig.~\ref{fig:results}(b)).
    }

}
\begin{table}[h]
        \centering
        \begin{tabular}{|c|c|c|c|}\hline
            & GT Training & Style Training & Style Adaptation \\
        \hline
        StyleImp & $\approx 12$ hrs & NA & $\ge 5$ hrs \\
        SNeRF & $\approx 12$ hrs & NA & 3-4 days \\
        StylizedNeRF & $\approx 12$ hrs & NA & $\ge 3$ hrs \\
        Ours & 20 mins & 20 mins & 45 secs \\
        \hline
        \end{tabular}
        \caption{\emph{Training Times:} We compare our training times with StyleImp~\cite{style_implicit_wacv}, SNeRF~\cite{snerf_siggraph}, StylizedNeRF~\cite{stylizednerf_cvpr}. Due to our compact representation, we get a training time of $\approx 20$ mins compared to at least $12$ hrs for others. The GT training and style training can be performed in parallel in our method. We disentangle the style training process from the style adaptation process which enables us to quickly adapt to any scene in under a minute. }
        \label{tab:time}
\end{table}
\vspace{-20pt}
\section{Conclusions}
{
    \label{sec:conc}

    In this paper, we presented StyleTRF, a compact and quick-to-optimize stylization technique which can generate stylized novel views of a scene. We have shown that our method can efficiently and faithfully incorporate style into a radiance field representation of a casually captured scene. We have qualitatively and quantitatively compared our method with the previous stylization methods. Our qualitative results and quantitative metrics demonstrate that StyleTRF is consistent across the views, having stylized the underlying 3D representation. We also reported the \emph{short and long-term} consistency metrics which are better in most cases compared to the present 3D stylization methods.    
    Concurrently, Zhang et al.\ \cite{arf_eccv} presented stylized novel view synthesis using voxel-based grids, partly similar to our method. But they chose PlenOxels~\cite{plenoxels} to represent the radiance fields. They focussed more on obtaining brush strokes while ours concentrates on geometric preserved fast-style adaptation.    
}

{
\noindent{\bf Acknowledgements: } 
 Sirikonda Dhawal was partially supported by DST. We also acknowledge support from the TCS Foundation through the Kohli
Center on Intelligent Systems (KCIS).
}
\bibliographystyle{ACM-Reference-Format}
\bibliography{main.bib}


\begin{thebibliography}{41}


\ifx \showCODEN    \undefined \def \showCODEN     #1{\unskip}     \fi
\ifx \showDOI      \undefined \def \showDOI       #1{#1}\fi
\ifx \showISBNx    \undefined \def \showISBNx     #1{\unskip}     \fi
\ifx \showISBNxiii \undefined \def \showISBNxiii  #1{\unskip}     \fi
\ifx \showISSN     \undefined \def \showISSN      #1{\unskip}     \fi
\ifx \showLCCN     \undefined \def \showLCCN      #1{\unskip}     \fi
\ifx \shownote     \undefined \def \shownote      #1{#1}          \fi
\ifx \showarticletitle \undefined \def \showarticletitle #1{#1}   \fi
\ifx \showURL      \undefined \def \showURL       {\relax}        \fi
\providecommand\bibfield[2]{#2}
\providecommand\bibinfo[2]{#2}
\providecommand\natexlab[1]{#1}
\providecommand\showeprint[2][]{arXiv:#2}

\bibitem[\protect\citeauthoryear{Aliev, Sevastopolsky, Kolos, Ulyanov, and
  Lempitsky}{Aliev et~al\mbox{.}}{2020}]%
        {npbg}
\bibfield{author}{\bibinfo{person}{Kara-Ali Aliev}, \bibinfo{person}{Artem
  Sevastopolsky}, \bibinfo{person}{Maria Kolos}, \bibinfo{person}{Dmitry
  Ulyanov}, {and} \bibinfo{person}{Victor Lempitsky}.}
  \bibinfo{year}{2020}\natexlab{}.
\newblock \showarticletitle{Neural Point-Based Graphics}. In
  \bibinfo{booktitle}{\emph{European Conference on Computer Vision (ECCV)}},
  \bibfield{editor}{\bibinfo{person}{Andrea Vedaldi}, \bibinfo{person}{Horst
  Bischof}, \bibinfo{person}{Thomas Brox}, {and} \bibinfo{person}{Jan-Michael
  Frahm}} (Eds.). \bibinfo{publisher}{Springer International Publishing},
  \bibinfo{address}{Cham}.
\newblock
\showISBNx{978-3-030-58542-6}


\bibitem[\protect\citeauthoryear{Barron, Mildenhall, Verbin, Srinivasan, and
  Hedman}{Barron et~al\mbox{.}}{2022}]%
        {mipnerf360}
\bibfield{author}{\bibinfo{person}{Jonathan~T. Barron}, \bibinfo{person}{Ben
  Mildenhall}, \bibinfo{person}{Dor Verbin}, \bibinfo{person}{Pratul~P.
  Srinivasan}, {and} \bibinfo{person}{Peter Hedman}.}
  \bibinfo{year}{2022}\natexlab{}.
\newblock \showarticletitle{Mip-NeRF 360: Unbounded Anti-Aliased Neural
  Radiance Fields}. In \bibinfo{booktitle}{\emph{Computer Vision and Pattern
  Recognition (CVPR)}}. \bibinfo{pages}{5470--5479}.
\newblock


\bibitem[\protect\citeauthoryear{Boss, Braun, Jampani, Barron, Liu, and
  Lensch}{Boss et~al\mbox{.}}{2020}]%
        {NeRD}
\bibfield{author}{\bibinfo{person}{Mark Boss}, \bibinfo{person}{Raphael Braun},
  \bibinfo{person}{Varun Jampani}, \bibinfo{person}{Jonathan~T. Barron},
  \bibinfo{person}{Ce Liu}, {and} \bibinfo{person}{Hendrik~P.A. Lensch}.}
  \bibinfo{year}{2020}\natexlab{}.
\newblock \showarticletitle{NeRD: Neural Reflectance Decomposition from Image
  Collections}. In \bibinfo{booktitle}{\emph{International Conference on
  Computer Vision (ICCV)}}.
\newblock


\bibitem[\protect\citeauthoryear{Boss, Jampani, Braun, Liu, Barron, and
  Lensch}{Boss et~al\mbox{.}}{2021}]%
        {neuralpil}
\bibfield{author}{\bibinfo{person}{Mark Boss}, \bibinfo{person}{Varun Jampani},
  \bibinfo{person}{Raphael Braun}, \bibinfo{person}{Ce Liu},
  \bibinfo{person}{Jonathan~T. Barron}, {and} \bibinfo{person}{Hendrik~P.A.
  Lensch}.} \bibinfo{year}{2021}\natexlab{}.
\newblock \showarticletitle{Neural-PIL: Neural Pre-Integrated Lighting for
  Reflectance Decomposition}. In \bibinfo{booktitle}{\emph{Advances in Neural
  Information Processing Systems (NeurIPS)}}.
\newblock


\bibitem[\protect\citeauthoryear{Cao, Wang, Nagao, and Nakamura}{Cao
  et~al\mbox{.}}{2020}]%
        {psnet_pointcloud_style}
\bibfield{author}{\bibinfo{person}{Xu Cao}, \bibinfo{person}{Weimin Wang},
  \bibinfo{person}{Katashi Nagao}, {and} \bibinfo{person}{Ryosuke Nakamura}.}
  \bibinfo{year}{2020}\natexlab{}.
\newblock \showarticletitle{PSNet: A Style Transfer Network for Point Cloud
  Stylization on Geometry and Color}. In \bibinfo{booktitle}{\emph{The IEEE
  Winter Conference on Applications of Computer Vision (WACV)}}.
\newblock


\bibitem[\protect\citeauthoryear{Chen, Xu, Geiger, Yu, and Su}{Chen
  et~al\mbox{.}}{2022}]%
        {tensorf}
\bibfield{author}{\bibinfo{person}{Anpei Chen}, \bibinfo{person}{Zexiang Xu},
  \bibinfo{person}{Andreas Geiger}, \bibinfo{person}{Jingyi Yu}, {and}
  \bibinfo{person}{Hao Su}.} \bibinfo{year}{2022}\natexlab{}.
\newblock \showarticletitle{TensoRF: Tensorial Radiance Fields}. In
  \bibinfo{booktitle}{\emph{European Conference on Computer Vision (ECCV)}}.
\newblock


\bibitem[\protect\citeauthoryear{Chiang, Tsai, Tseng, Lai, and Chiu}{Chiang
  et~al\mbox{.}}{2022}]%
        {style_implicit_wacv}
\bibfield{author}{\bibinfo{person}{Pei-Ze Chiang}, \bibinfo{person}{Meng-Shiun
  Tsai}, \bibinfo{person}{Hung-Yu Tseng}, \bibinfo{person}{Wei-Sheng Lai},
  {and} \bibinfo{person}{Wei-Chen Chiu}.} \bibinfo{year}{2022}\natexlab{}.
\newblock \showarticletitle{Stylizing 3D Scene via Implicit Representation and
  HyperNetwork}. In \bibinfo{booktitle}{\emph{Winter Conference on Applications
  of Computer Vision (WACV)}}.
\newblock


\bibitem[\protect\citeauthoryear{De~Lathauwer}{De~Lathauwer}{2008}]%
        {btd_paper}
\bibfield{author}{\bibinfo{person}{Lieven De~Lathauwer}.}
  \bibinfo{year}{2008}\natexlab{}.
\newblock \showarticletitle{Decompositions of a Higher-Order Tensor in Block
  Terms—Part II: Definitions and Uniqueness}.
\newblock \bibinfo{journal}{\emph{SIAM J. Matrix Anal. Appl.}}
  \bibinfo{volume}{30}, \bibinfo{number}{3} (\bibinfo{year}{2008}),
  \bibinfo{pages}{1033--1066}.
\newblock
\urldef\tempurl%
\url{https://doi.org/10.1137/070690729}
\showDOI{\tempurl}
\showeprint{https://doi.org/10.1137/070690729}


\bibitem[\protect\citeauthoryear{{Fridovich-Keil and Yu}, Tancik, Chen, Recht,
  and Kanazawa}{{Fridovich-Keil and Yu} et~al\mbox{.}}{2022}]%
        {plenoxels}
\bibfield{author}{\bibinfo{person}{{Fridovich-Keil and Yu}},
  \bibinfo{person}{Matthew Tancik}, \bibinfo{person}{Qinhong Chen},
  \bibinfo{person}{Benjamin Recht}, {and} \bibinfo{person}{Angjoo Kanazawa}.}
  \bibinfo{year}{2022}\natexlab{}.
\newblock \showarticletitle{Plenoxels: Radiance Fields without Neural
  Networks}. In \bibinfo{booktitle}{\emph{Computer Vision and Pattern
  Recognition (CVPR)}}.
\newblock


\bibitem[\protect\citeauthoryear{Gatys, Ecker, and Bethge}{Gatys
  et~al\mbox{.}}{2016}]%
        {gatys}
\bibfield{author}{\bibinfo{person}{Leon~A. Gatys},
  \bibinfo{person}{Alexander~S. Ecker}, {and} \bibinfo{person}{Matthias
  Bethge}.} \bibinfo{year}{2016}\natexlab{}.
\newblock \showarticletitle{Image Style Transfer Using Convolutional Neural
  Networks}. In \bibinfo{booktitle}{\emph{2016 IEEE Conference on Computer
  Vision and Pattern Recognition (CVPR)}}. \bibinfo{pages}{2414--2423}.
\newblock
\urldef\tempurl%
\url{https://doi.org/10.1109/CVPR.2016.265}
\showDOI{\tempurl}


\bibitem[\protect\citeauthoryear{Gera, Dastjerdi, Renaud, Narayanan, and
  Lalonde}{Gera et~al\mbox{.}}{2022}]%
        {hdr_gera}
\bibfield{author}{\bibinfo{person}{Pulkit Gera}, \bibinfo{person}{Mohammad
  Reza~Karimi Dastjerdi}, \bibinfo{person}{Charles Renaud},
  \bibinfo{person}{P.~J. Narayanan}, {and} \bibinfo{person}{Jean-François
  Lalonde}.} \bibinfo{year}{2022}\natexlab{}.
\newblock \showarticletitle{Casual Indoor HDR Radiance Capture from
  Omnidirectional Images}. In \bibinfo{booktitle}{\emph{British Machine Vision
  Conference (BMVC)}}.
\newblock


\bibitem[\protect\citeauthoryear{Giampouras, Rontogiannis, and
  Kofidis}{Giampouras et~al\mbox{.}}{2022}]%
        {btd_tsp}
\bibfield{author}{\bibinfo{person}{Paris~V. Giampouras},
  \bibinfo{person}{Athanasios~A. Rontogiannis}, {and}
  \bibinfo{person}{Eleftherios Kofidis}.} \bibinfo{year}{2022}\natexlab{}.
\newblock \showarticletitle{Block-Term Tensor Decomposition Model Selection and
  Computation: The Bayesian Way}.
\newblock \bibinfo{journal}{\emph{IEEE Transactions on Signal Processing}}
  \bibinfo{volume}{70} (\bibinfo{year}{2022}), \bibinfo{pages}{1704--1717}.
\newblock
\urldef\tempurl%
\url{https://doi.org/10.1109/TSP.2022.3159029}
\showDOI{\tempurl}


\bibitem[\protect\citeauthoryear{Gortler, Grzeszczuk, Szeliski, and
  Cohen}{Gortler et~al\mbox{.}}{1996}]%
        {lumigraph}
\bibfield{author}{\bibinfo{person}{Steven~J. Gortler}, \bibinfo{person}{Radek
  Grzeszczuk}, \bibinfo{person}{Richard Szeliski}, {and}
  \bibinfo{person}{Michael~F. Cohen}.} \bibinfo{year}{1996}\natexlab{}.
\newblock \showarticletitle{The Lumigraph}. In
  \bibinfo{booktitle}{\emph{Proceedings of the 23rd Annual Conference on
  Computer Graphics and Interactive Techniques}}
  \emph{(\bibinfo{series}{SIGGRAPH '96})}. \bibinfo{publisher}{Association for
  Computing Machinery}, \bibinfo{address}{New York, NY, USA}.
\newblock
\showISBNx{0897917464}
\urldef\tempurl%
\url{https://doi.org/10.1145/237170.237200}
\showDOI{\tempurl}


\bibitem[\protect\citeauthoryear{Ha, Dai, and Le}{Ha et~al\mbox{.}}{2017}]%
        {hypernetworks}
\bibfield{author}{\bibinfo{person}{David Ha}, \bibinfo{person}{Andrew Dai},
  {and} \bibinfo{person}{Quoc~V. Le}.} \bibinfo{year}{2017}\natexlab{}.
\newblock \showarticletitle{HyperNetworks}. In
  \bibinfo{booktitle}{\emph{International Conference on Learning
  Representations (ICLR)}}.
\newblock


\bibitem[\protect\citeauthoryear{Hedman, Srinivasan, Mildenhall, Barron, and
  Debevec}{Hedman et~al\mbox{.}}{2021}]%
        {bakednerf}
\bibfield{author}{\bibinfo{person}{Peter Hedman}, \bibinfo{person}{Pratul~P.
  Srinivasan}, \bibinfo{person}{Ben Mildenhall}, \bibinfo{person}{Jonathan~T.
  Barron}, {and} \bibinfo{person}{Paul Debevec}.}
  \bibinfo{year}{2021}\natexlab{}.
\newblock \showarticletitle{Baking Neural Radiance Fields for Real-Time View
  Synthesis}. In \bibinfo{booktitle}{\emph{International Conference on Computer
  Vision (ICCV)}}.
\newblock


\bibitem[\protect\citeauthoryear{Huang, Wang, Luo, Ma, Jiang, Zhu, Li, and
  Liu}{Huang et~al\mbox{.}}{2017}]%
        {real_time_nst_video}
\bibfield{author}{\bibinfo{person}{Haozhi Huang}, \bibinfo{person}{Hao Wang},
  \bibinfo{person}{Wenhan Luo}, \bibinfo{person}{Lin Ma},
  \bibinfo{person}{Wenhao Jiang}, \bibinfo{person}{Xiaolong Zhu},
  \bibinfo{person}{Zhifeng Li}, {and} \bibinfo{person}{Wei Liu}.}
  \bibinfo{year}{2017}\natexlab{}.
\newblock \showarticletitle{Real-Time Neural Style Transfer for Videos}. In
  \bibinfo{booktitle}{\emph{Computer Vision and Pattern Recognition (CVPR)}}.
\newblock


\bibitem[\protect\citeauthoryear{Huang, Tseng, Saini, Singh, and Yang}{Huang
  et~al\mbox{.}}{2021}]%
        {style_view_point_cloud}
\bibfield{author}{\bibinfo{person}{Hsin-Ping Huang}, \bibinfo{person}{Hung-Yu
  Tseng}, \bibinfo{person}{Saurabh Saini}, \bibinfo{person}{Maneesh Singh},
  {and} \bibinfo{person}{Ming-Hsuan Yang}.} \bibinfo{year}{2021}\natexlab{}.
\newblock \showarticletitle{Learning to Stylize Novel Views}. In
  \bibinfo{booktitle}{\emph{International Conference on Computer Vision
  (ICCV)}}.
\newblock


\bibitem[\protect\citeauthoryear{Huang and Belongie}{Huang and
  Belongie}{2017}]%
        {adain}
\bibfield{author}{\bibinfo{person}{Xun Huang} {and} \bibinfo{person}{Serge
  Belongie}.} \bibinfo{year}{2017}\natexlab{}.
\newblock \showarticletitle{Arbitrary Style Transfer in Real-Time with Adaptive
  Instance Normalization}. In \bibinfo{booktitle}{\emph{International
  Conference on Computer Vision (ICCV)}}.
\newblock


\bibitem[\protect\citeauthoryear{Huang, He, Yuan, Lai, and Gao}{Huang
  et~al\mbox{.}}{2022}]%
        {stylizednerf_cvpr}
\bibfield{author}{\bibinfo{person}{Yi-Hua Huang}, \bibinfo{person}{Yue He},
  \bibinfo{person}{Yu-Jie Yuan}, \bibinfo{person}{Yu-Kun Lai}, {and}
  \bibinfo{person}{Lin Gao}.} \bibinfo{year}{2022}\natexlab{}.
\newblock \showarticletitle{StylizedNeRF: Consistent 3D Scene Stylization as
  Stylized NeRF via 2D-3D Mutual Learning}. In
  \bibinfo{booktitle}{\emph{Computer Vision and Pattern Recognition (CVPR)}}.
\newblock


\bibitem[\protect\citeauthoryear{Johnson, Alahi, and Fei-Fei}{Johnson
  et~al\mbox{.}}{2016}]%
        {johnson}
\bibfield{author}{\bibinfo{person}{Justin Johnson}, \bibinfo{person}{Alexandre
  Alahi}, {and} \bibinfo{person}{Li Fei-Fei}.} \bibinfo{year}{2016}\natexlab{}.
\newblock \showarticletitle{Perceptual Losses for Real-Time Style Transfer and
  Super-Resolution}. In \bibinfo{booktitle}{\emph{European Conference on
  Computer Vision (ECCV)}}.
\newblock


\bibitem[\protect\citeauthoryear{Kellnhofer, Jebe, Jones, Spicer, Pulli, and
  Wetzstein}{Kellnhofer et~al\mbox{.}}{2021}]%
        {neural_lumigraph}
\bibfield{author}{\bibinfo{person}{Petr Kellnhofer}, \bibinfo{person}{Lars~C.
  Jebe}, \bibinfo{person}{Andrew Jones}, \bibinfo{person}{Ryan Spicer},
  \bibinfo{person}{Kari Pulli}, {and} \bibinfo{person}{Gordon Wetzstein}.}
  \bibinfo{year}{2021}\natexlab{}.
\newblock \showarticletitle{Neural Lumigraph Rendering}. In
  \bibinfo{booktitle}{\emph{Computer Vision and Pattern Recognition (CVPR)}}.
\newblock


\bibitem[\protect\citeauthoryear{Kingma and Ba}{Kingma and Ba}{2014}]%
        {adam}
\bibfield{author}{\bibinfo{person}{Diederik~P. Kingma} {and}
  \bibinfo{person}{Jimmy Ba}.} \bibinfo{year}{2014}\natexlab{}.
\newblock \bibinfo{title}{Adam: A Method for Stochastic Optimization}.
\newblock
\newblock
\urldef\tempurl%
\url{https://doi.org/10.48550/ARXIV.1412.6980}
\showDOI{\tempurl}


\bibitem[\protect\citeauthoryear{Li, Fang, Yang, Wang, Lu, and Yang}{Li
  et~al\mbox{.}}{2017}]%
        {image_style_wct}
\bibfield{author}{\bibinfo{person}{Yijun Li}, \bibinfo{person}{Chen Fang},
  \bibinfo{person}{Jimei Yang}, \bibinfo{person}{Zhaowen Wang},
  \bibinfo{person}{Xin Lu}, {and} \bibinfo{person}{Ming-Hsuan Yang}.}
  \bibinfo{year}{2017}\natexlab{}.
\newblock \showarticletitle{Universal Style Transfer via Feature Transforms}.
  In \bibinfo{booktitle}{\emph{Advances in Neural Information Processing
  Systems}}, \bibfield{editor}{\bibinfo{person}{I.~Guyon},
  \bibinfo{person}{U.~Von Luxburg}, \bibinfo{person}{S.~Bengio},
  \bibinfo{person}{H.~Wallach}, \bibinfo{person}{R.~Fergus},
  \bibinfo{person}{S.~Vishwanathan}, {and} \bibinfo{person}{R.~Garnett}}
  (Eds.), Vol.~\bibinfo{volume}{30}. \bibinfo{publisher}{Curran Associates,
  Inc.}
\newblock


\bibitem[\protect\citeauthoryear{Lin, Maire, Belongie, Hays, Perona, Ramanan,
  Doll{\'a}r, and Zitnick}{Lin et~al\mbox{.}}{2014}]%
        {coco}
\bibfield{author}{\bibinfo{person}{Tsung-Yi Lin}, \bibinfo{person}{Michael
  Maire}, \bibinfo{person}{Serge Belongie}, \bibinfo{person}{James Hays},
  \bibinfo{person}{Pietro Perona}, \bibinfo{person}{Deva Ramanan},
  \bibinfo{person}{Piotr Doll{\'a}r}, {and} \bibinfo{person}{C.~Lawrence
  Zitnick}.} \bibinfo{year}{2014}\natexlab{}.
\newblock \showarticletitle{Microsoft COCO: Common Objects in Context}. In
  \bibinfo{booktitle}{\emph{European Conference on Computer Vision (ECCV)}},
  \bibfield{editor}{\bibinfo{person}{David Fleet}, \bibinfo{person}{Tomas
  Pajdla}, \bibinfo{person}{Bernt Schiele}, {and} \bibinfo{person}{Tinne
  Tuytelaars}} (Eds.).
\newblock


\bibitem[\protect\citeauthoryear{Liu, Lin, He, Li, Wang, Li, Sun, Li, and
  Ding}{Liu et~al\mbox{.}}{2021}]%
        {video_adaattn}
\bibfield{author}{\bibinfo{person}{Songhua Liu}, \bibinfo{person}{Tianwei Lin},
  \bibinfo{person}{Dongliang He}, \bibinfo{person}{Fu Li},
  \bibinfo{person}{Meiling Wang}, \bibinfo{person}{Xin Li},
  \bibinfo{person}{Zhengxing Sun}, \bibinfo{person}{Qian Li}, {and}
  \bibinfo{person}{Errui Ding}.} \bibinfo{year}{2021}\natexlab{}.
\newblock \showarticletitle{AdaAttN: Revisit Attention Mechanism in Arbitrary
  Neural Style Transfer}.
\newblock \bibinfo{journal}{\emph{International Conference on Computer Vision
  (ICCV)}}.
\newblock


\bibitem[\protect\citeauthoryear{Mildenhall, Hedman, Martin-Brualla,
  Srinivasan, and Barron}{Mildenhall et~al\mbox{.}}{2022}]%
        {rawnerf}
\bibfield{author}{\bibinfo{person}{Ben Mildenhall}, \bibinfo{person}{Peter
  Hedman}, \bibinfo{person}{Ricardo Martin-Brualla}, \bibinfo{person}{Pratul~P.
  Srinivasan}, {and} \bibinfo{person}{Jonathan~T. Barron}.}
  \bibinfo{year}{2022}\natexlab{}.
\newblock \showarticletitle{NeRF in the Dark: High Dynamic Range View Synthesis
  From Noisy Raw Images}. In \bibinfo{booktitle}{\emph{Computer Vision and
  Pattern Recognition (CVPR)}}.
\newblock


\bibitem[\protect\citeauthoryear{Mildenhall, Srinivasan, Ortiz-Cayon,
  Kalantari, Ramamoorthi, Ng, and Kar}{Mildenhall et~al\mbox{.}}{2019}]%
        {mildenhall2019llff}
\bibfield{author}{\bibinfo{person}{Ben Mildenhall}, \bibinfo{person}{Pratul~P.
  Srinivasan}, \bibinfo{person}{Rodrigo Ortiz-Cayon},
  \bibinfo{person}{Nima~Khademi Kalantari}, \bibinfo{person}{Ravi Ramamoorthi},
  \bibinfo{person}{Ren Ng}, {and} \bibinfo{person}{Abhishek Kar}.}
  \bibinfo{year}{2019}\natexlab{}.
\newblock \showarticletitle{Local Light Field Fusion: Practical View Synthesis
  with Prescriptive Sampling Guidelines}.
\newblock \bibinfo{journal}{\emph{ACM Transactions on Graphics (TOG)}}
  (\bibinfo{year}{2019}).
\newblock


\bibitem[\protect\citeauthoryear{Mildenhall, Srinivasan, Tancik, Barron,
  Ramamoorthi, and Ng}{Mildenhall et~al\mbox{.}}{2020}]%
        {nerf}
\bibfield{author}{\bibinfo{person}{Ben Mildenhall}, \bibinfo{person}{Pratul~P.
  Srinivasan}, \bibinfo{person}{Matthew Tancik}, \bibinfo{person}{Jonathan~T.
  Barron}, \bibinfo{person}{Ravi Ramamoorthi}, {and} \bibinfo{person}{Ren Ng}.}
  \bibinfo{year}{2020}\natexlab{}.
\newblock \showarticletitle{NeRF: Representing Scenes as Neural Radiance Fields
  for View Synthesis}. In \bibinfo{booktitle}{\emph{European Conference on
  Computer Vision (ECCV)}}.
\newblock


\bibitem[\protect\citeauthoryear{Nguyen-Phuoc, Liu, and Xiao}{Nguyen-Phuoc
  et~al\mbox{.}}{2022}]%
        {snerf_siggraph}
\bibfield{author}{\bibinfo{person}{Thu Nguyen-Phuoc}, \bibinfo{person}{Feng
  Liu}, {and} \bibinfo{person}{Lei Xiao}.} \bibinfo{year}{2022}\natexlab{}.
\newblock \showarticletitle{SNeRF: Stylized Neural Implicit Representations for
  3D Scenes}.
\newblock \bibinfo{journal}{\emph{ACM Trans. Graph.}} \bibinfo{volume}{41},
  \bibinfo{number}{4}, Article \bibinfo{articleno}{142} (\bibinfo{date}{jul}
  \bibinfo{year}{2022}), \bibinfo{numpages}{11}~pages.
\newblock
\showISSN{0730-0301}
\urldef\tempurl%
\url{https://doi.org/10.1145/3528223.3530107}
\showDOI{\tempurl}


\bibitem[\protect\citeauthoryear{Reiser, Peng, Liao, and Geiger}{Reiser
  et~al\mbox{.}}{2021}]%
        {kilonerf}
\bibfield{author}{\bibinfo{person}{Christian Reiser}, \bibinfo{person}{Songyou
  Peng}, \bibinfo{person}{Yiyi Liao}, {and} \bibinfo{person}{Andreas Geiger}.}
  \bibinfo{year}{2021}\natexlab{}.
\newblock \showarticletitle{KiloNeRF: Speeding up Neural Radiance Fields with
  Thousands of Tiny MLPs}. In \bibinfo{booktitle}{\emph{International
  Conference on Computer Vision (ICCV)}}.
\newblock


\bibitem[\protect\citeauthoryear{Ruder, Dosovitskiy, and Brox}{Ruder
  et~al\mbox{.}}{2016}]%
        {ruder}
\bibfield{author}{\bibinfo{person}{Manuel Ruder}, \bibinfo{person}{Alexey
  Dosovitskiy}, {and} \bibinfo{person}{Thomas Brox}.}
  \bibinfo{year}{2016}\natexlab{}.
\newblock \showarticletitle{Artistic Style Transfer for Videos}. In
  \bibinfo{booktitle}{\emph{German Conference on Pattern Recognition (GCPR)}},
  \bibfield{editor}{\bibinfo{person}{Bodo Rosenhahn} {and}
  \bibinfo{person}{Bjoern Andres}} (Eds.).
\newblock


\bibitem[\protect\citeauthoryear{Sch\"{o}nberger and Frahm}{Sch\"{o}nberger and
  Frahm}{2016}]%
        {colmap}
\bibfield{author}{\bibinfo{person}{Johannes~Lutz Sch\"{o}nberger} {and}
  \bibinfo{person}{Jan-Michael Frahm}.} \bibinfo{year}{2016}\natexlab{}.
\newblock \showarticletitle{Structure-from-Motion Revisited}. In
  \bibinfo{booktitle}{\emph{Computer Vision and Pattern Recognition (CVPR)}}.
\newblock


\bibitem[\protect\citeauthoryear{Sheng, Lin, Shao, and Wang}{Sheng
  et~al\mbox{.}}{2018}]%
        {avatar}
\bibfield{author}{\bibinfo{person}{Lu Sheng}, \bibinfo{person}{Ziyi Lin},
  \bibinfo{person}{Jing Shao}, {and} \bibinfo{person}{Xiaogang Wang}.}
  \bibinfo{year}{2018}\natexlab{}.
\newblock \showarticletitle{Avatar-Net: Multi-scale Zero-Shot Style Transfer by
  Feature Decoration}. In \bibinfo{booktitle}{\emph{Computer Vision and Pattern
  Recognition (CVPR)}}.
\newblock


\bibitem[\protect\citeauthoryear{Srinivasan, Deng, Zhang, Tancik, Mildenhall,
  and Barron}{Srinivasan et~al\mbox{.}}{2021}]%
        {nerv2020}
\bibfield{author}{\bibinfo{person}{Pratul~P. Srinivasan},
  \bibinfo{person}{Boyang Deng}, \bibinfo{person}{Xiuming Zhang},
  \bibinfo{person}{Matthew Tancik}, \bibinfo{person}{Ben Mildenhall}, {and}
  \bibinfo{person}{Jonathan~T. Barron}.} \bibinfo{year}{2021}\natexlab{}.
\newblock \showarticletitle{NeRV: Neural Reflectance and Visibility Fields for
  Relighting and View Synthesis}. In \bibinfo{booktitle}{\emph{Computer Vision
  and Pattern Recognition (CVPR)}}.
\newblock


\bibitem[\protect\citeauthoryear{Svoboda, Anoosheh, Osendorfer, and
  Masci}{Svoboda et~al\mbox{.}}{2020}]%
        {image_style_tspr}
\bibfield{author}{\bibinfo{person}{Jan Svoboda}, \bibinfo{person}{Asha
  Anoosheh}, \bibinfo{person}{Christian Osendorfer}, {and}
  \bibinfo{person}{Jonathan Masci}.} \bibinfo{year}{2020}\natexlab{}.
\newblock \showarticletitle{Two-Stage Peer-Regularized Feature Recombination
  for Arbitrary Image Style Transfer}. In \bibinfo{booktitle}{\emph{Computer
  Vision and Pattern Recognition (CVPR)}}.
\newblock


\bibitem[\protect\citeauthoryear{Teed and Deng}{Teed and Deng}{2021}]%
        {raft}
\bibfield{author}{\bibinfo{person}{Zachary Teed} {and} \bibinfo{person}{Jia
  Deng}.} \bibinfo{year}{2021}\natexlab{}.
\newblock \showarticletitle{RAFT: Recurrent All-Pairs Field Transforms for
  Optical Flow (Extended Abstract)}. In \bibinfo{booktitle}{\emph{Proceedings
  of the Thirtieth International Joint Conference on Artificial Intelligence,
  {IJCAI-21}}}, \bibfield{editor}{\bibinfo{person}{Zhi-Hua Zhou}} (Ed.).
  \bibinfo{publisher}{International Joint Conferences on Artificial
  Intelligence Organization}.
\newblock


\bibitem[\protect\citeauthoryear{Tucker and Snavely}{Tucker and
  Snavely}{2020}]%
        {multi_plane_images}
\bibfield{author}{\bibinfo{person}{Richard Tucker} {and} \bibinfo{person}{Noah
  Snavely}.} \bibinfo{year}{2020}\natexlab{}.
\newblock \showarticletitle{Single-View View Synthesis With Multiplane Images}.
  In \bibinfo{booktitle}{\emph{Computer Vision and Pattern Recognition
  (CVPR)}}.
\newblock


\bibitem[\protect\citeauthoryear{Wang, Yang, Xu, and Liu}{Wang
  et~al\mbox{.}}{2020}]%
        {ReReVST2020}
\bibfield{author}{\bibinfo{person}{Wenjing Wang}, \bibinfo{person}{Shuai Yang},
  \bibinfo{person}{Jizheng Xu}, {and} \bibinfo{person}{Jiaying Liu}.}
  \bibinfo{year}{2020}\natexlab{}.
\newblock \showarticletitle{Consistent Video Style Transfer via Relaxation and
  Regularization}.
\newblock \bibinfo{journal}{\emph{{IEEE} Trans. Image Process.}}
  (\bibinfo{year}{2020}).
\newblock


\bibitem[\protect\citeauthoryear{Zhang, Kolkin, Bi, Luan, Xu, Shechtman, and
  Snavely}{Zhang et~al\mbox{.}}{2022}]%
        {arf_eccv}
\bibfield{author}{\bibinfo{person}{Kai Zhang}, \bibinfo{person}{Nick Kolkin},
  \bibinfo{person}{Sai Bi}, \bibinfo{person}{Fujun Luan},
  \bibinfo{person}{Zexiang Xu}, \bibinfo{person}{Eli Shechtman}, {and}
  \bibinfo{person}{Noah Snavely}.} \bibinfo{year}{2022}\natexlab{}.
\newblock \showarticletitle{ARF: Artistic Radiance Fields}. In
  \bibinfo{booktitle}{\emph{European Conference on Computer Vision (ECCV)}}.
\newblock


\bibitem[\protect\citeauthoryear{Zhang, Luan, Wang, Bala, and Snavely}{Zhang
  et~al\mbox{.}}{2021}]%
        {physg2020}
\bibfield{author}{\bibinfo{person}{Kai Zhang}, \bibinfo{person}{Fujun Luan},
  \bibinfo{person}{Qianqian Wang}, \bibinfo{person}{Kavita Bala}, {and}
  \bibinfo{person}{Noah Snavely}.} \bibinfo{year}{2021}\natexlab{}.
\newblock \showarticletitle{PhySG: Inverse Rendering with Spherical Gaussians
  for Physics-based Material Editing and Relighting}. In
  \bibinfo{booktitle}{\emph{Computer Vision and Pattern Recognition (CVPR)}}.
\newblock


\bibitem[\protect\citeauthoryear{Zhou, Hu, Lin, Guo, and Shum}{Zhou
  et~al\mbox{.}}{2005}]%
        {source_radiance_fields}
\bibfield{author}{\bibinfo{person}{Kun Zhou}, \bibinfo{person}{Yaohua Hu},
  \bibinfo{person}{Stephen Lin}, \bibinfo{person}{Baining Guo}, {and}
  \bibinfo{person}{Heung-Yeung Shum}.} \bibinfo{year}{2005}\natexlab{}.
\newblock \showarticletitle{Precomputed Shadow Fields for Dynamic Scenes}. In
  \bibinfo{booktitle}{\emph{ACM SIGGRAPH 2005 Papers}} (Los Angeles,
  California) \emph{(\bibinfo{series}{SIGGRAPH '05})}.
  \bibinfo{publisher}{Association for Computing Machinery},
  \bibinfo{address}{New York, NY, USA}, \bibinfo{pages}{1196–1201}.
\newblock
\showISBNx{9781450378253}
\urldef\tempurl%
\url{https://doi.org/10.1145/1186822.1073332}
\showDOI{\tempurl}


\end{thebibliography}
\end{document}